\documentclass{article}

\usepackage[final]{packages/corl_2026} 
\usepackage{graphicx}  

\newif\ifanonymous
\anonymoustrue 

\DeclareRobustCommand{\anon}[1]{%
  \ifanonymous
  \else
    #1%
  \fi
}

\title{EgoHTR: Egocentric 4D Demonstrations of Human Terrain Traversal}

\author{
  Alex Brandes$^{1}$ \And
  Haig Conti Georges Sajelian$^{4}$ \And 
  Manthan Patel$^{1}$ \And 
  Dominik Hollidt$^{1}$ \And 
  Chenhao Li$^{1}$ \And 
  Matthias Heyrman$^{1}$ \And
  Oliver Hausdörfer$^{4}$ \And 
  Manuel Kaufmann$^{1}$ \And 
  Xi Wang$^{1}$ \And 
  Jonas Frey$^{2,3}$ \And 
  Angela P. Schoellig$^{4}$ \And
  Christian Holz$^{1}$ \And 
  Marc Pollefeys$^{1}$ \And 
  Marco Hutter$^{1}$ \AND
  \textbf{$^{1}$ETH Zurich \quad $^{2}$Stanford University \quad $^{3}$UC Berkeley \quad $^{4}$TU Munich}
}

\usepackage[utf8]{inputenc}
\usepackage[OT1]{fontenc}

\usepackage{a4}

\usepackage{fancyhdr}

\usepackage{textcomp}\usepackage{gensymb}

\usepackage[english]{isodate}

\usepackage{multicol}

\usepackage{graphicx}

\usepackage{subfigure}

\usepackage{booktabs}
\usepackage{array}
\usepackage{multirow}
\usepackage{enumitem}

\usepackage{color}
\usepackage{colortbl}
\definecolor{black}{rgb}{0,0,0}
\definecolor{white}{rgb}{1,1,1}
\definecolor{darkred}{rgb}{0.5,0,0}
\definecolor{darkgreen}{rgb}{0,0.5,0}
\definecolor{darkblue}{rgb}{0,0,0.5}

\usepackage{amsmath}
\usepackage{amssymb}

\usepackage{nicefrac}

\usepackage{upgreek}

\usepackage{isomath}

\usepackage{rotating}

\usepackage{pdfpages}
\includepdfset{pages={-}, frame=true, pagecommand={\thispagestyle{fancy}}}

\usepackage{pgfgantt}

\PassOptionsToPackage{hyphens}{url}

\usepackage{cleveref}
\let\cref\Cref

\usepackage{tabularx}
\newcolumntype{Y}{>{\centering\arraybackslash}X}

\usepackage{makecell} 
\usepackage{siunitx}
\usepackage{multirow}
\usepackage{float}
\usepackage{fontawesome5} 
\usepackage{xcolor}       
\usepackage{booktabs}
\usepackage{ragged2e}
\usepackage{amssymb}
\usepackage{geometry}
\usepackage{caption}
\usepackage{siunitx}
\usepackage{threeparttable}
\usepackage{tikz}
\usetikzlibrary{arrows.meta, positioning}
\definecolor{Gray}{gray}{0.95}

\newcommand{\cm}{\checkmark} 
\newcommand{\smicon}[1]{\scalebox{0.75}{#1}}

\renewcommand{\anon}[1]{#1}

\begin{document}
\addtocontents{toc}{\protect\setcounter{tocdepth}{-1}}
\maketitle


\begin{figure}[htbp]
    \vspace{-10pt}
    \centering
    \includegraphics[trim={0.1cm 0.1cm 0.1cm 0.1cm}, clip, width=\textwidth]{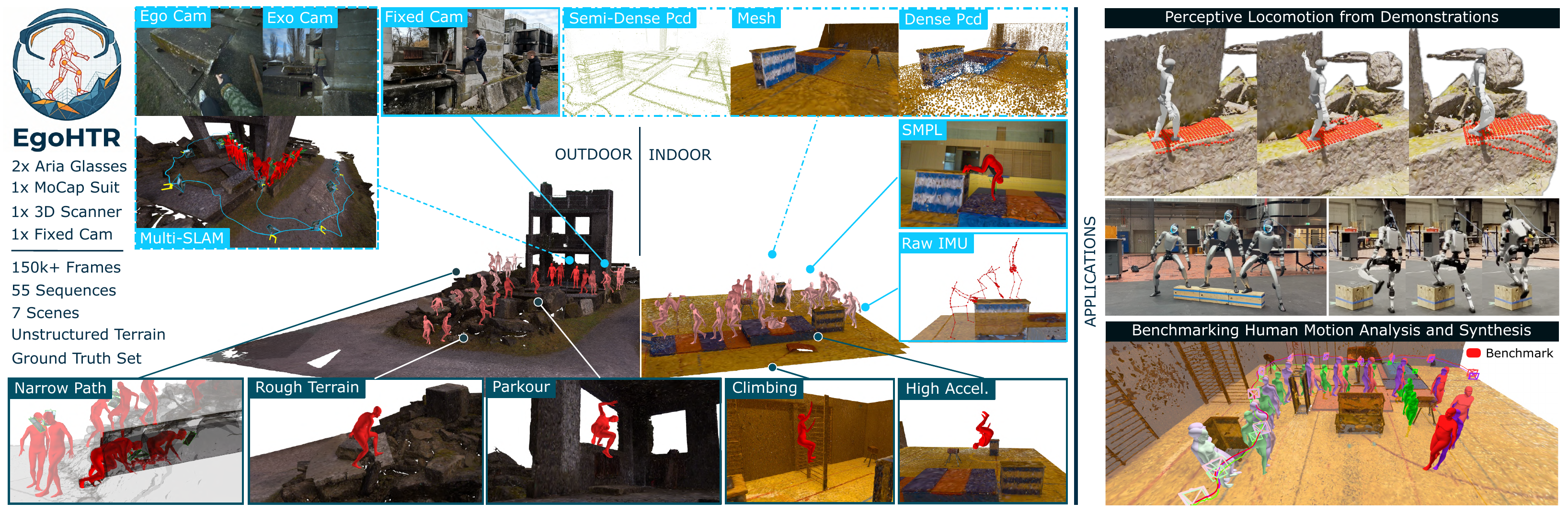}
    \captionsetup{labelfont=bf, font=small, skip=5pt, justification=justified}
    \caption{EgoHTR dataset preview, a set of in-the-wild 4D human-scene demonstrations focused on rough terrain. Left: Example outdoor (debris field) and indoor (gym hall) scene, highlighting the reconstructed 4D human-scene sequences with its provided modalities (light blue; ego- and exocentric Aria video and SLAM streams, 3D mesh and point cloud, parametrized model, raw IMU) and the diversity of human motions (dark blue). Right: Downstream applications verifying the utility of the dataset, including the successful deployment of perceptive locomotion policies.}
    \label{fig:cover}
    \vspace{-10pt}
\end{figure}

\begin{abstract} 
Deploying humanoid robots in unstructured terrain remains an open problem. 
While classic reinforcement learning struggles with the sheer complexity of real-world interactions, more promising methods leveraging human priors remain limited to models lacking contextual awareness.  
The restricted motion synthesis is a direct consequence of existing dataset pipelines failing to capture human-scene sequences in challenging environments. 
To bridge this gap between humanoid learning and scene reconstruction, we introduce the Egocentric Human-Terrain Reconstruction (EgoHTR) dataset. 
We develop and open-source a reconstruction pipeline capturing 55 scene-aligned 4D human motion sequences in diverse, complex environments using a multi-sensor setup of egocentric wearables and a portable 3D scanner.
The resulting dataset comprises over 150k frames, which we evaluate against motion-capture ground truth, demonstrating state-of-the-art accuracy and establishing a rigorous benchmark for human motion analysis and synthesis.
Further, we leverage this data to train perceptive locomotion policies, demonstrating hardware deployment on a Unitree G1 for reconstructed reference motions. 
Our pipeline enables community-driven dataset extensions and factors the problem to help researchers build foundational, context-aware robots that reliably traverse uneven terrain.  \anon{\url{https://egohtr.github.io}}
\end{abstract}

\section{Introduction}
\label{sec:introduction}

Terrain traversal with humanoid robots is challenging due to its dynamic gait and vast exploration space induced by their high degrees of freedom.
While perceptive reinforcement learning-based (RL) policies conditioned on depth images or elevation maps have achieved success in traversing rough environments \citep{zhang2026rpl, rudin2025parkour, sun2025dpl, Long2024LearningHL}, they typically depend on task-specific, complex reward engineering. 
Learning from human behavior has emerged as a promising alternative demonstrating robust performances \citep{fu2024humanplus, yin2025unitrackerlearninguniversalwholebody, liao2025beyondmimicmotiontrackingversatile, chen2025gmt, zhao2025resmimicgeneralmotiontracking, zhang2025trackmotionsdisturbances, luo2025sonicsupersizingmotiontracking, videomimic, zhang2026meshmimicgeometryawarehumanoidmotion, wang2026embodmocapinthewild4dhumanscene}. 
By mimicking human reference data, these methods provide strong kinematic priors that efficiently constrain the RL search space, establishing a scalable foundation for learning diverse, whole-body locomotion skills.

However, the learning and deploying of mimic-based, context-aware controllers for humanoid embodiments in-the-wild is limited by the scarcity of current training datasets. 
In particular, existing sources lack multi-modal recordings of diverse motions in rough terrain. 
Available datasets (\cref{tab:dataset_comparison_final}) either miss dense scene information \citep{AMASS:ICCV:2019, harvey2020lafan1, Guo2022HML3D, kaufmann2023emdb, grauman2024egoexo4dunderstandingskilledhuman, zhang2025motion, ma2024nymeriamassivecollectionmultimodal}, are limited to spatially constrained setups \citep{PROX2019, RICHHuang:CVPR:2022}, focus exclusively on indoor environments \citep{zhang2022egobody, zheng2022gimo, li2023object, kim2024parahome, detone2026nymeriaplusenrichingnymeriadataset}, or feature flat and structured terrains \citep{Guzov2021HPS, Dai_2023_CVPR}. 
Although recent works attempt to address this gap using vision-based human-scene reconstruction \citep{videomimic, zhang2026meshmimicgeometryawarehumanoidmotion, wang2026embodmocapinthewild4dhumanscene, liu2026joint, chen2025human3r}, they suffer from perceptual failures and limited accuracy due to the model's reliance on exocentric modalities and on context-limited training datasets. This data gap highlights the inherent difficulty of in-the-wild collection, which typically requires a compromise between scalability and reconstruction accuracy. Consequently, there is a need for an open-source capture pipeline that balances these competing factors.  

To address this, our key idea is to construct an extendable sensor configuration by combining the benefits of commercially available sensors, including a wearable motion-capture suit (spatial independence), an egocentric multi-sensor headset (state-of-the-art SLAM), and a 3D scanner (dense scene context). 
This approach overcomes these bottlenecks, enabling centimeter-precision 4D reconstruction of dynamic trajectories, ranging from ladder climbing to backflips, in truly unstructured settings (e.g., collapsed buildings, debris fields). 
Using this recording pipeline, we introduce the Egocentric Human-Terrain Reconstruction (EgoHTR) dataset, summarized in \cref{fig:cover}. 
EgoHTR features 7 unstructured scenes, 8 subjects, and 55 diverse multi-modal sequences focused on rough terrain traversal, totaling \SI{150}{k} frames with an average sequence length of \SI{90}{\second}. 
Beyond benchmarking 4D reconstruction methods, EgoHTR provides a high-fidelity foundation for training and deploying context-aware locomotion policies.       
Through this framework, our work makes three key contributions.
(1) Reconstruction Pipeline: We release a scalable 4D human-scene reconstruction pipeline to enable community-driven dataset extensions.
(2) Multi-Modal Dataset: We introduce a dataset featuring multi-modal human motion sequences grounded within high-resolution rough 3D scene meshes.
(3) Applications: We establish a comprehensive benchmark for evaluating human pose and shape (HPS) estimation and 4D human-scene reconstruction on rough terrain. Additionally, by providing the centimeter-precise reference motions proven strictly necessary for foothold-critical tasks, the dataset serves as a foundational resource for training context-aware locomotion policies where current methods fail.


\section{Related Work}
\label{sec:relatedwork}

\paragraph{Human Motion Datasets}
\Cref{tab:dataset_comparison_final} compares recent human motion datasets. Early marker-based and text-conditioned databases provide high-fidelity kinematics but lack publicly available scene context \citep{AMASS:ICCV:2019, Guo2022HML3D, Black_CVPR_2023, bones2026aidatasets}. To incorporate scene interactions, exocentric and multi-view approaches enabled 4D human-scene analysis, yet they remain spatially constrained to indoor or structured environments \citep{PROX2019, zhang2022egobody, grauman2024egoexo4dunderstandingskilledhuman, RICHHuang:CVPR:2022}. For in-the-wild scaling, recent methods leverage wearable and egocentric sensors to fuse motion capture with 3D environments. However, they still suffer from missing sensory modalities \citep{kaufmann2023emdb, ma2024nymeriamassivecollectionmultimodal} and from environmental biases toward flat \citep{kaufmann2023emdb, zheng2022gimo}, confined \citep{RICHHuang:CVPR:2022} or structured spaces \citep{Guzov2021HPS, Dai_2023_CVPR}. Consequently, there remains a critical data gap for multi-modal human motion in unstructured, rough terrain.

\paragraph{Learning Locomotion from Human Demonstrations}
Research on humanoid locomotion policies learned from human demonstrations has progressed
from scene-agnostic motion tracking toward perceptive imitation that grounds
motion in 3D scene geometry. Initial efforts established high-fidelity tracking
of MoCap references without exteroceptive perception
\citep{yin2025unitrackerlearninguniversalwholebody,
liao2025beyondmimicmotiontrackingversatile, chen2025gmt,
zhao2025resmimicgeneralmotiontracking, zhang2025trackmotionsdisturbances,
zhang2025hub, wang2025experts, luo2025sonicsupersizingmotiontracking}, along
with shadowing \citep{fu2024humanplus}, teleoperation \citep{ze2025twist, luo2025sonicsupersizingmotiontracking} and pure retargeting frameworks
\citep{yang2025omniretarget, ze2025gmr}. 
More recent perceptive approaches condition policies on scene geometry \citep{wu2026perceptivehumanoidparkourchaining, videomimic, zhang2026meshmimicgeometryawarehumanoidmotion, wang2026embodmocapinthewild4dhumanscene}. However, their reliance on monocular reconstructions \citep{videomimic, zhang2026meshmimicgeometryawarehumanoidmotion} or primitive obstacles \citep{wu2026perceptivehumanoidparkourchaining, zhuang2026deepwholebodyparkour} restricts learning to coarse environments. Because these methods supervise general joint tracking rather than explicit end-effector alignment, they fail to provide the contact precision required for complex topography. EgoHTR addresses this by providing in-the-wild data with scene-aligned motions, enabling learned policies to move beyond flat ground and discrete obstacles.

\begin{table}[t]
    \vspace{-20pt}
    \centering
    \captionsetup{labelfont=bf, font=small, skip=5pt, justification=justified}
    \caption{Comparison of Human Motion Datasets.}
    \label{tab:dataset_comparison_final}
    
    \begin{minipage}[c]{0.80\linewidth}
        \scriptsize
        \setlength{\tabcolsep}{3pt} 
        \renewcommand{\arraystretch}{0.9} 
        
        \resizebox{\linewidth}{!}{
        \setlength{\aboverulesep}{0pt}
        \setlength{\belowrulesep}{2pt}
        \begin{tabular}{l l c c c c c c c c c c c c}
        \toprule
            \textbf{Dataset} & \textbf{Year} & \multicolumn{3}{c}{\textbf{Scale}} & \multicolumn{6}{c}{\textbf{Modalities}} & \multicolumn{3}{c}{\textbf{Location}} \\
        \cmidrule(lr){3-5} \cmidrule(lr){6-11} \cmidrule(lr){12-14}
            & & \smicon{\faClock} & \smicon{\faImages} & \smicon{\faHourglassHalf} & 
            \smicon{\faMale} & \smicon{\faGlasses} &  \smicon{\faBinoculars} &  \smicon{\faVideo} & \smicon{\faEye} & \smicon{\faTree} & 
            \smicon{\faHome} & \smicon{\faCloudSun} & \smicon{\faMountain} \\
        \midrule
            \rowcolor{Gray}
            3DPW \citep{vonMarcard2018} & 2018 & 0.5 & 0.05 & 0.48 & \cm & \cm & \cm & & & & & \cm & \\
            AMASS \citep{AMASS:ICCV:2019} & 2019 & 42 & 0.9 & 0.22 & \cm & & & & & & \cm & & \\
            \rowcolor{Gray}
            PROX \citep{PROX2019} & 2019 & 0.92 & 0.1 & 4.6 & \cm & & & \cm & & \cm & \cm & & \\  
            LaFAN1 \citep{harvey2020lafan1} & 2020 & 4.6 & 0.5 & 3.6 & \cm & & & \cm & & & \cm & & \\
            \rowcolor{Gray}
            HPS \citep{Guzov2021HPS} & 2021 & 4.5 & 0.3 & 8.2 & \cm & \cm & & & & \cm & \cm & \cm & \\
            RICH \citep{RICHHuang:CVPR:2022} & 2022 & 5.34 & 0.58 & 2.26 & \cm & & & \cm & & \cm & \cm & \cm \\
            \rowcolor{Gray}
            GIMO \citep{zheng2022gimo} & 2022 & 1.2 & 0.125 & 0.32 & \cm & \cm & & & \cm & \cm & \cm & & \\
            BEDLAM \citep{Black_CVPR_2023} & 2023 & 14.8 & 1.6 & 0.38 & \cm & & & & & \cm & \cm & \cm \\
            \rowcolor{Gray}
            EMDB \citep{kaufmann2023emdb} & 2023 & 0.97 & 0.11 & 0.72 & \cm & \cm & \cm & & & & \cm & \cm & \\
            SLOPER4D \citep{Dai_2023_CVPR} & 2023 & 1.38 & 0.1 & 5.5 & \cm & \cm & \cm & & & \cm & & \cm & \\
            \rowcolor{Gray}
            EgoExo4D \citep{grauman2024egoexo4dunderstandingskilledhuman} & 2024 & 1286 & 14 & 1.42 & & \cm & & \cm & \cm & $(\cm)$ & \cm & \cm & \cm \\
            NymeriaPlus \citep{detone2026nymeriaplusenrichingnymeriadataset} & 2026 & 300 & 260 & 15 & \cm & \cm & \cm & & \cm & $(\cm)$ & \cm & \cm & \\
            \rowcolor{Gray}
            BONES-SEED \citep{bones2026aidatasets} & 2026 & 288 & 124.5 & 0.12 & \cm & & & & & & \cm & & \\
            \textbf{EgoHTR (Ours)} & 2026 & 1.37 & 0.15 & 1.5 & \cm & \cm & \cm & \cm & \cm & \cm & \cm & \cm & \cm \\
        \bottomrule
        \end{tabular}
        }
    \end{minipage}
    \hfill
    \begin{minipage}[c]{0.18\linewidth}
        \scriptsize
        \renewcommand{\arraystretch}{1.0}
        \resizebox{\linewidth}{!}{
        \begin{tabular}{@{}c l@{}}
            \multicolumn{2}{@{}l}{\textbf{Scale}} \\
            \faClock & Hours \\
            \faImages & Frames (M) \\
            \faHourglassHalf & Avg. Seq. (min) \\
            \addlinespace
            \multicolumn{2}{@{}l}{\textbf{Modalities}} \\
            \faMale & Param. Body \\
            \faGlasses & Subject (ego) \\
            \faBinoculars & 2nd Person \\
            \faVideo & Fixed Cam. \\
            \faEye & Gaze \\
            \faTree & Scene Mesh\\
            \addlinespace
            \multicolumn{2}{@{}l}{\textbf{Location}} \\
            \faHome & Indoor \\
            \faCloudSun & Outdoor \\
            \faMountain & Rough Terrain \\
            \addlinespace
            \multicolumn{2}{@{}p{0.9\linewidth}@{}}{$(\cm)$ limited scene reconstruction} \\
        \end{tabular}
        }
    \end{minipage}
    \vspace{-15pt}
\end{table}

\paragraph{Data-Driven Human Mesh Recovery}
While isolated human mesh recovery (HMR) has seen significant progress across diverse modalities \citep{sarandi2024neural, multi-hmr2024, wham:cvpr:2024, li2023ego, PIPCVPR2022}, research is now shifting toward 4D human-scene reconstruction. Current scene-aware methods generally fall into two categories. The first relies on independent modules for human and scene geometry, necessitating complex, post-hoc optimization for alignment \citep{videomimic, zhang2026meshmimicgeometryawarehumanoidmotion, wang2026embodmocapinthewild4dhumanscene, liu2026joint}. The second attempts to learn 4D reconstruction implicitly using unified vision models \citep{chen2025human3r, sur2026unicon3rcontactaware3dhumanscene} or terrain-aware IMU posers \citep{jiang2022transformer}. However, these unified approaches suffer heavily from the human-scene data gap, struggling to produce physically plausible real-world interactions.
A comprehensive review of the related works is provided in \cref{app:extended_related_work}.
\section{Data Acquisition Pipeline} \label{sec:methodology}

\begin{figure}[htbp]
    \vspace{-10pt}
    \centering
    \includegraphics[trim={0.0cm 0.0cm 0.0cm 0.0cm}, clip, width=\textwidth]{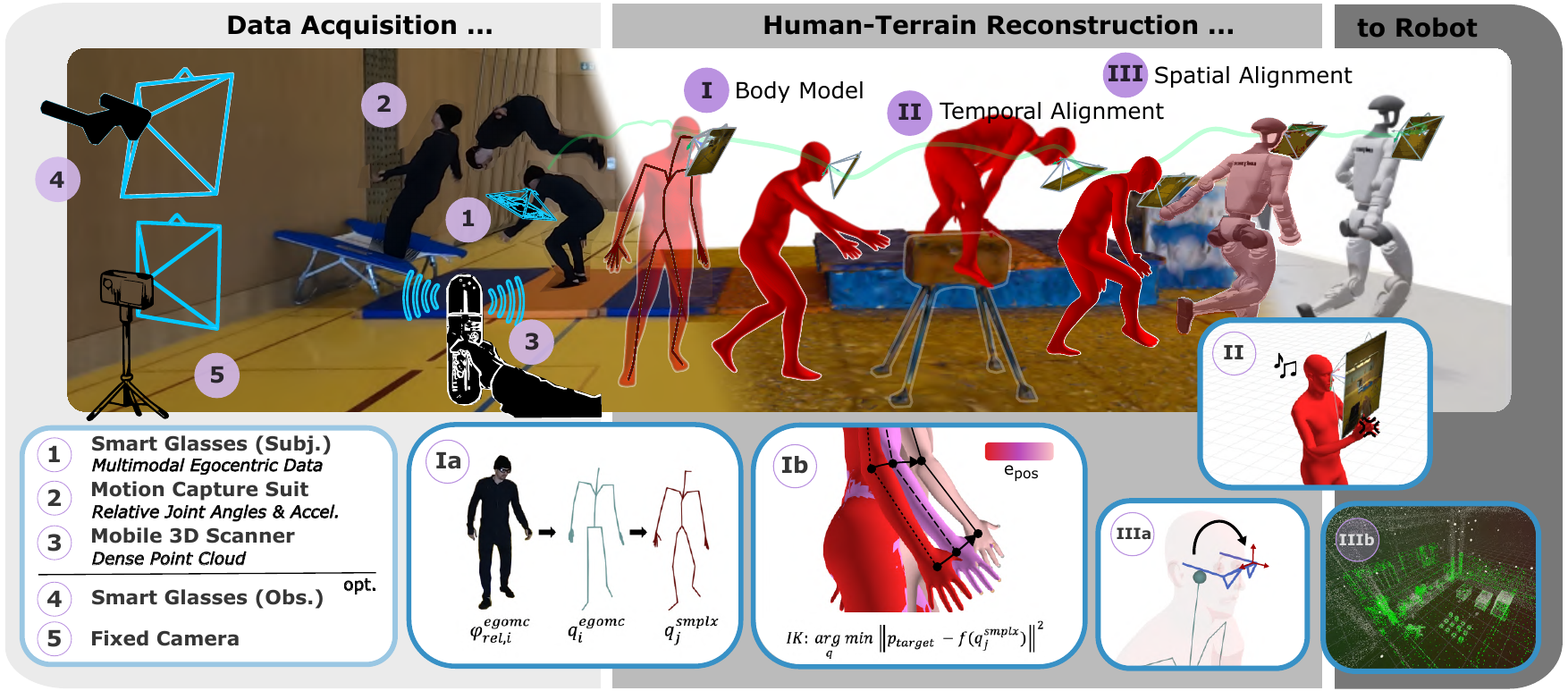}
    \captionsetup{labelfont=bf, font=small, skip=5pt, justification=justified}
    \caption{Overview of the implemented dataset generation pipeline. We categorize the methodology into two stages: data acquisition and human-terrain reconstruction -- enabling human2robot retargeting (\cref{sec:Human2Robot}). A capture system with 3 core sensors (Aria glasses, MoCap suit, 3D Scanner) provide data for a three stage human-scene reconstruction: (I) body model parametrization, (II) temporal and (III) spatial alignment.}
    \label{fig:method}
    \vspace{-10pt}
\end{figure}

\subsection{Capture System} \label{sec:capture system}
Our goal is the recording of human motion data and its environment mesh from wearable sensors in arbitrary scenarios, achieving an optimal trade-off between deployment usability and reconstruction accuracy. 
Our methodology, shown in \cref{fig:method}, is based on a multi-modal sensor suite consisting of: (1) a head-mounted multi-sensor Project Aria glasses Gen.~1 \citep{engel2023projectaria}, (2) an IMU-based Rokoko Pro II MoCap suit, and (3) a high-resolution Leica BLK2GO 3D scene scanner. 
To capture additional perspectives, this setup can optionally include (4) a second observer-worn Project Aria glasses and (5) a static fixed-view camera. 
This setup offers a rich set of sensory streams, enabling the reconstruction of multi-modal human-scene sequences. Detailed data specifications and further hardware details are listed in the \cref{app:sec_method}. 
Unlike traditional lab-grade setups, our full sensor suite is portable and designed to capture data in any unconstrained, real-world environment.

\subsection{Human-Scene Reconstruction} \label{sec:humanterrain generation}
As visualized in \cref{fig:method}, our 4D human-scene reconstruction follows three stages: (I) a per-frame parameterization of human pose and shape, (II) temporal synchronization of all sensor streams, (III) spatial alignment in a common world frame. Extended information is given in \cref{app:sec_methodlogy}.

\paragraph{Body Model (I)}
We parameterize the human subject using the SMPL-X model \citep{SMPL-X:2019}. Because our motion capture suit provides kinematics for only the 22 primary body joints \citep{SMPL:2015}, we fix hand and facial parameters to the identity pose, providing a modular baseline for future extensions. Since body shapes ($\boldsymbol{\beta}$) are assumed constant within a sequence, we estimate only the articulated pose $\boldsymbol{\theta}$ and the global translation $\boldsymbol{t}$. Shape parameters are obtained using state-of-the-art toolkits \citep{Black_CVPR_2023, SMPL-X:2019}.

Our retargeting pipeline transfers the MoCap suit skeleton to the SMPL-X topology in two stages. First, a rotational retargeting step (\cref{fig:method}, Ia) converts the source global quaternions into target relative rotation matrices, applying a per-joint offset to compensate for systematic skeletal axis misalignment and structural differences, following the work of \citet{ze2025gmr}. Second, to mitigate joint position errors introduced by structural differences between the skeletons, we refine the pose via an optimization-based inverse kinematics (IK) step (\cref{fig:method}, Ib). This minimizes the 3D positional discrepancy between the limb joints of the SMPL-X and the corresponding MoCap joint positions.



\paragraph{Temporal Alignment (II)} To ensure temporal synchronization, we register the data streams of the MoCap suit and the Aria glasses using a distinct hand clap at the beginning of every sequence \citep{Guzov2021HPS}. By omitting restrictive, hardwired Pulse-Per-Second (PPS) synchronization cables, we trade sub-millisecond lab precision for the unconstrained mobility required to capture in-the-wild human locomotion. While the microsecond-level resolution of the audio stream (\SI{48}{kHz}) bounds the alignment error strictly to the IMU frequency (\SI{100}{Hz}), empirical evaluations demonstrate a cross-sensor alignment of less than \SI{60}{ms}. By restricting our recorded sequences to a maximum length of \SI{5}{min} (as recommended by Rokoko), we constrain hardware clock drift to negligible levels, eliminating the need for active, continuous time synchronization. (see \cref{fig:temporal_alignment}) 

\paragraph{Spatial Alignment (III)}
We perform a two-stage spatial alignment to register the human motion sequences into the global coordinate frame of the 3D scene. First, we anchor the body model kinematics to the closed-loop trajectory provided by the Aria glasses \citep{engel2023projectaria}, by defining the relationship between the Aria camera $C_A$ and the MoCap suit head joint $H$. Under the assumption that the glasses do not physically move relative to the head, we model the relationship as a static translation offset (denoted ${}_{\mathcal{C_A}}r_{C_{AH}}$). Unlike \citet{ma2024nymeriamassivecollectionmultimodal}, this approach enables us to eliminate global IMU drift by relying entirely on Aria SLAM (providing ground truth accuracy as investigated by \citet{Krishnan_2025_ICCV}). However, it necessitates an empirical calibration of the offset.
Consequently, we can express the body sequence $P$ in the Aria world frame $\mathcal{W}_A$,
\begin{equation} \label{eq:pose_aria}
    T_{\mathcal{W}_AP} = T_{\mathcal{W}_AC_A} ~ T_{C_AH} ~ T_{HP}; \quad 
    T_{C_AH} = \begin{bmatrix} I & {}_\mathcal{C_A}r_{C_AH} \\ \mathbf{0} & 1 \end{bmatrix}
\end{equation}

Second, to align the reconstructed human sequence with the 3D scanned scene (S), we estimate the rigid transform between the Aria world frame $\mathcal{W}_A$, and the world frame $\mathcal{W} = \mathcal{W_S}$. We initialize a coarse mapping $\tilde{T}_{\mathcal{W}\mathcal{W}_A}$ using the Visual Grounded Geometry Transformer (VGGT) \citep{wang2025vggt} to estimate the relative transform $f_{\text{T}}$ between the initial scanner $I_{C_{S,0}}$ and Aria glasses images $I_{C_{A,t}}$. This alignment is refined via Iterative Closest Point (ICP), registering the semi-dense Aria point cloud $\mathcal{P}_A$ against the dense environmental scan $\mathcal{P}_S$ to yield the final transform $T_{\mathcal{W}P}$:
\begin{equation} \label{eq:world_align}
\begin{aligned} 
\tilde{T}_{\mathcal{W}\mathcal{W}_A} &= T_{\mathcal{W}C_{S,0}} ~ f_{\text{T}}(I_{C_{S,0}}, I_{C_{A,t}}) ~ (T_{\mathcal{W}_AC_{A,t}})^{-1} \\
T_{\mathcal{W}P} &= T_{\mathcal{W}\mathcal{W}_A} ~ T_{\mathcal{W}_AP} = f_{\text{ICP}}(\tilde{T}_{\mathcal{W}\mathcal{W}_A}, \mathcal{P}_S, \mathcal{P}_A) ~ T_{\mathcal{W}_AP}
\end{aligned}
\end{equation}

By leveraging the complementary modalities of the sensor suite, our framework enables the capture of diverse motions across arbitrary environments. Therefore, this methodology overcomes the limitations of spatial constraints and optimization capabilities introduced by existing pipelines \citep{PROX2019, Guzov2021HPS, Dai_2023_CVPR, RICHHuang:CVPR:2022}.  

\section{EgoHTR Dataset Overview}
\label{sec:overview_dataset}

\subsection{Dataset Statistics}
We introduce the Egocentric Human-Terrain Reconstruction (EgoHTR) dataset. 
The dataset features 8 subjects (4 male, 4 female) navigating through 7 diverse indoor and outdoor scenes, as illustrated in \cref{fig:scenes}. 
These environments span explorable areas from \SI{25}{m^2} to \SI{1000}{m^2}, chosen to cover agile motions, including parkour, flips, crawling, jumping, bar balancing, and narrow path crossing. Settings range from controlled laboratory parkour (70--\SI{140}{cm} boxes, 10--\SI{50}{cm} stepping stones) to unstructured in-the-wild scenes such as an office room, a cluttered gym hall, and an outdoor debris field for rough terrain navigation among collapsed buildings and subterranean pipes.
In total, EgoHTR consists of 55 continuous human-terrain interaction sequences, yielding \SI{1.37}{hours} of parameterized (see \cref{sec:gt_acquisition}) motion data, resulting in approximately \SI{150}{k} synchronized frames captured at \SI{30}{fps} (detailed in Table~\ref{tab:dataset}). 
To support multi-view and observer-centric motion analysis, 36 of these sequences (totaling \SI{0.88}{hours}) include synchronized data from an external observer wearing a secondary pair of Aria glasses. 
An external fixed-camera image stream is provided for the majority of the sequences. Furthermore, as described in \cref{sec:gt_acquisition}, we provide a dedicated marker-based motion capture ground truth test set containing \SI{0.7}{hours} under lab-conditions. This subset enables rigorous validation of our reconstruction accuracy (discussed in \cref{par:quant_eval}) prior to large-scale, in-the-wild data collection.

\subsection{Dataset Evaluation}
In this section, we address the following questions: How competitive is EgoHTR's local and global pose estimation accuracy compared to existing human-centric datasets? Can the proposed methodology maintain accurate spatial and temporal reconstruction under in-the-wild conditions? 
\paragraph{Quantitative Evaluation} \label{par:quant_eval}
Following standard evaluation metrics in recent literature \citep{PROX2019, RICHHuang:CVPR:2022, Dai_2023_CVPR, wang2025prompthmrpromptablehumanmesh, multi-hmr2024, wang2024tram, wham:cvpr:2024, liu2026joint, chen2025human3r}, as detailed in \cref{sec:app_quantitative_dataset_eval}, we evaluate the local and global HPS estimation on our test set and compare the results against reported state-of-the-art baselines, given in \cref{tab:baseline_comp}.  
In terms of local pose accuracy, EgoHTR reports an MPJPE (mean per joint position error) of \SI{73.2}{mm} and a PA-MPJPE (procrustes-aligned MPJPE) of \SI{54.3}{mm}. 
Compared to the strongest baseline, SLOPER4D \citep{Dai_2023_CVPR}, EgoHTR achieves a 6.2\% lower MPJPE and a 2.0\% lower PA-MPJPE. Crucially, this performance improvement is achieved under substantially harder conditions. While SLOPER4D is restricted to basic walking and stair climbing in structured urban terrain, EgoHTR captures highly dynamic motions (e.g. backflips, parkour, crawling) across severe rough terrain. 
Furthermore, SLOPER4D as well as PROX \citep{PROX2019} and RICH \citep{RICHHuang:CVPR:2022} suffer from high failure rates under occlusion or dynamic motions due to their exocentric dependencies, while HPS \citep{Guzov2021HPS} lacks scalability to rough terrain. 
EgoHTR overcomes these bottlenecks through robust egocentric sensing. To prove a robust reconstruction quality outside of laboratory conditions, \cref{fig:itw_vs_eval} reports the foot-floating ratio (FFR) and foot sliding (FS) metrics
introduced by \citet{liu2026joint} across the full dataset. They are split between in-the-wild and MoCap-validated scenes and demonstrate comparable fidelity. The overall metric scores further underline the temporal and spatial robustness of our human-scene alignment.

While exhibiting higher errors than human-only datasets \citep{vonMarcard2018, kaufmann2023emdb} due to the added complexity of environmental interactions, EgoHTR sets a new state-of-the-art for 4D human-scene reconstruction.
In addition, to our knowledge, EgoHTR is the first human motion dataset to report global HPS estimates, currently used in recent 4D human-scene reconstruction work \citep{liu2026joint, chen2025human3r} (world and world-aligned MPJPE, root translation error (RTE)).

\begin{figure}[h]
    \vspace{-10pt}  
    \centering
    \includegraphics[trim={0cm 0cm 0cm 0cm}, clip, width=\textwidth]{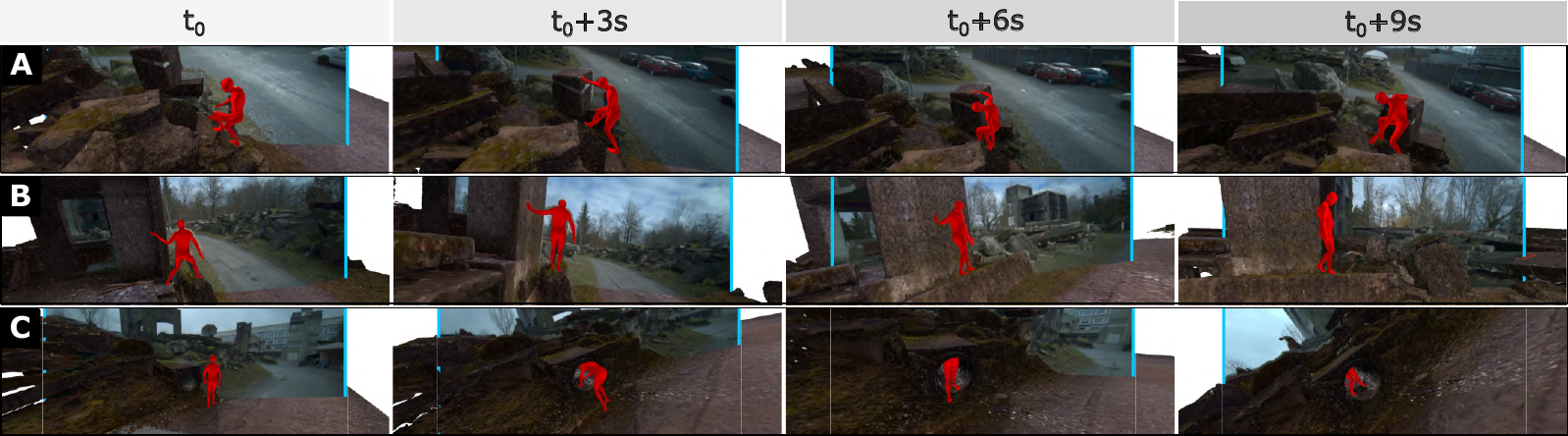}
    \captionsetup{labelfont=bf, font=small, skip=5pt, justification=justified}
    \caption{The image streams show the aligned 3D environment (background) with the estimated 3D human mesh (red) overlaid with the projected observer image (blue border) of versatile motions: walking through a debris field (A), balancing over a narrow beam (B), and crawling into a pipe (C).}
    \label{fig:quali_itw}
    \vspace{-10pt}
\end{figure}

\paragraph{Qualitative Evaluation}
EgoHTR enables accurate and physically plausible human motion reconstruction in challenging in-the-wild environments, as shown in \cref{fig:quali_itw}. 
In contrast to existing approaches, EgoHTR reliably reconstructs human motion on rough and uneven terrain while limiting common reconstruction artifacts such as foot skating, ground penetration, and temporally inconsistent body motion. 
Through the integration of Aria glasses, the reconstructed body poses are accurately localized within the global scene, enabling temporally stable human-scene alignment over long motion sequences. 
Importantly, EgoHTR remains robust under highly challenging conditions, including rough terrain, severe self-occlusions, rapid motion, and uncommon body configurations where purely vision-based approaches often become unstable or fail. 
As illustrated in \cref{fig:quali_mesh_recovery}, EgoHTR furthermore enables accurate pose reconstruction in confined environments such as narrow pipes and boxes, where visibility is heavily restricted.
Notably, we observe no meaningful qualitative degradation between controlled indoor motion-capture recordings and in-the-wild outdoor sequences, indicating strong robustness and domain transfer capabilities of the proposed multi-sensor reconstruction pipeline.

\begin{table*}[h]
    \vspace{-0pt}
    \centering
    \captionsetup{labelfont=bf, font=small, skip=5pt, justification=justified}

    \begin{minipage}[t]{0.48\textwidth}
        \centering
        \renewcommand{\arraystretch}{1.0}

        \caption{Comparison of EgoHTR to human-only (\faUser) and human-scene (\faUser \thinspace \faTree) reconstruction dataset baselines. \faImages ~ assigns the number of frames (in thousands). Best results (w/o human-only) are in \textbf{bold}, second-best \underline{underlined}.}

        \resizebox{\linewidth}{!}{%
            \setlength{\aboverulesep}{1pt}
            \setlength{\belowrulesep}{2pt}
            \begin{tabular}{l c cc c}
                \toprule
                \textbf{Baselines} & \faImages & MPJPE $\downarrow$ & PA-MPJPE $\downarrow$ & Type \\
                \midrule
                \rowcolor{gray!10}
                3DPW '18 & 29.3 & -- & 26 & \faUser\\
                \rowcolor{gray!10}
                EMDB '23 & 58.3 & -- & 23 & \faUser\\
                PROX '19 & 0.18 & 167.1 & 72.0 & \faUser \thinspace \faTree\\
                HPS '21 & -- & 93.1 & -- & \faUser \thinspace \faTree\\
                RICH '22 & 125 & 161.8 & 63.7 & \faUser \thinspace \faTree\\
                SLOPER4D '23 & 20 & \underline{78.01} & \underline{55.4} & \faUser \thinspace \faTree\\ 
                \midrule
                \textbf{Ours (local)} & 72.6 & \textbf{73.2} & \textbf{54.3} & \faUser \thinspace \faTree \\
                \midrule
                \midrule
                & \multicolumn{1}{c}{} & W-MPJPE $\downarrow$ & WA-MPJPE $\downarrow$ & RTE (\%) $\downarrow$ \\
                \midrule
                \textbf{Ours (global)} & 72.6 & \textbf{151.3} & \textbf{66.7} & \textbf{0.09} \\
                \bottomrule
            \end{tabular}%
        }

        \label{tab:baseline_comp}
    \end{minipage}
    \hfill
    \begin{minipage}[t]{0.5\textwidth}
        \centering
        \vspace{-0pt}

        \captionof{figure}{Fidelity metrics evaluated on in-the-wild (blue) and test subset (brown) motions of EgoHTR.}
        \label{fig:itw_vs_eval}

        \includegraphics[
            width=\linewidth,
            height=0.24\textheight,
            keepaspectratio
        ]{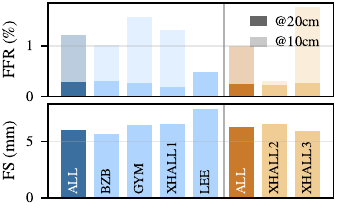}
    \end{minipage}%

\end{table*}

\section{Applications} \label{sec:applications}
\subsection{Perceptive Locomotion from Demonstrations}
\label{sec:perceptive_wbc}

\begin{figure}[htbp]
    \centering
    \includegraphics[trim={0cm 0cm 0cm 0cm}, clip, width=\textwidth]{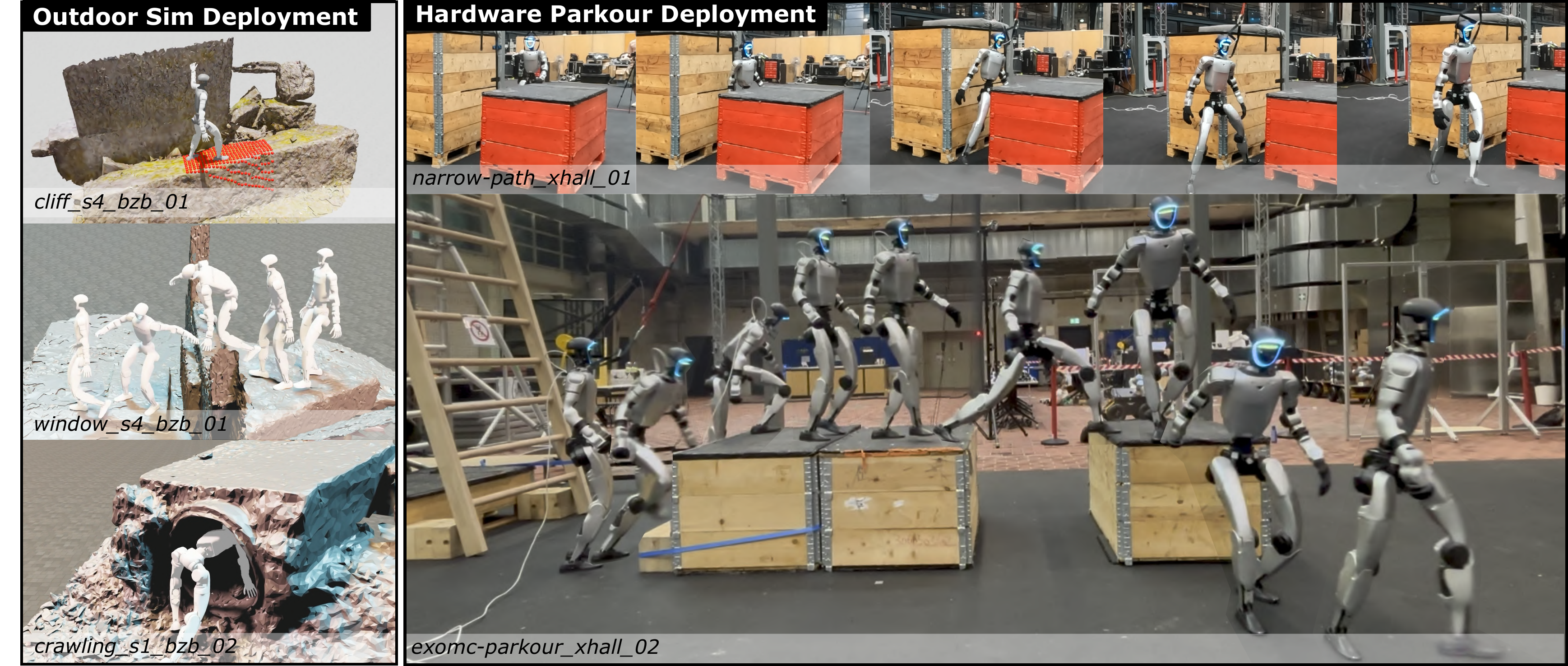}
    \captionsetup{labelfont=bf, font=small, skip=5pt, justification=justified}
    \caption{Demonstrating perceptive whole-body control policies trained on in-the-wild EgoHTR reference motions and deployed in outdoor simulation (left) and parkour real world environments (right).}
    \label{fig:deployment}
    \vspace{-10pt}
\end{figure}

Training whole-body mimic policies on sparse, complex terrain requires reference motions that specify both \emph{where} the foot lands and \emph{when} it bears load. EgoHTR provides terrain-aligned reference motions together with per-frame foot-contact annotations on non-primitive terrain.
In simulation we train a Unitree G1 humanoid robot to track the provided human motion reference. Per reference-clip we train a separate expert policy using PPO. The actor observes proprioception, the retargeted reference joint command, and a yaw-aligned terrain height scan. The critic additionally receives privileged body states and ankle contact wrenches. Beyond standard mimic rewards on body pose and end-effector position, we introduce a temporal foot-contact reward. It is critical on sparse terrain, where distance-only tracking might lead to degenerate solutions where the foot hovers without bearing load. The ablation in \cref{tab:footlock_ablation} shows that adding the contact reward yields faster convergence and a higher final success rate, especially on sparse terrain (stepping stones, shown in \cref{app:stepping_stones}). Episodes are initialized via adaptive reference state initialization restricted to interpenetration-free frames. Further training details and deployment randomization are provided in \cref{app:training_pipeline}. Hardware deployment of trained motion policies is shown in \cref{fig:deployment} and the corresponding Real2Sim2Real pipeline specifications are listed in \cref{tab:pipeline_timeline}.

\paragraph{Reference Motion Accuracy}
This raises a question central to the dataset's justification: could one obtain the same training signal more cheaply from monocular human-scene reconstruction? Such pipelines \citep{videomimic, zhang2026meshmimicgeometryawarehumanoidmotion} recover scene geometry but exhibit global estimation errors on rough terrain that typically exceed \SI{0.1}{m}. Whether errors of this magnitude are absorbed by retargeting and policy learning, or whether they preclude training, has remained an open question. It determines whether high-precision capture pipelines like ours are necessary or merely convenient. \newline
To quantify the required precision, we inject Gaussian noise into the reference root position and evaluate policy training on two contrasting terrains: stepping stones (sparse, foothold-critical) and beam (dense contact, single discrete obstacle). 
\Cref{fig:precision} reveals a consistent two-regime behavior across both environments: training tolerates noise up to roughly $\sigma = \SI{0.05}{m}$ and collapses above \SI{0.1}{m}. Two conclusions follow: First, retargeting and policy learning together absorb non-trivial misalignment, so perfect data is not required. The \SI{0.05}{m} tolerance window is consistent across motion classes, suggesting it reflects a property of the learning problem. 
Second, global drift in current monocular methods yields pose (W- and WA-MPJPE) and root translation errors (RTE) that exceed this narrow tolerance window. In contrast, EgoHTR prevents drift by anchoring the body model to the Aria SLAM trajectory and dense scan (Eq.\ref{eq:world_align}). This approach meets the centimeter-precise threshold, enabling perceptive policy training where scalable alternatives fail. We conclude that foothold-precise locomotion strictly requires human-terrain coupling.

\begin{table*}[t]
    \vspace{-0pt}
    \centering
    \captionsetup{labelfont=bf, font=small, skip=5pt, justification=justified}

    \begin{minipage}[t]{0.50\textwidth}
        \centering
        \renewcommand{\arraystretch}{1.0}

        \caption{Foot-contact reward impact analysis on EgoHTR atomic motions. We derive the binary foothold information from the motion capture suit. $\pi_0$: world-frame tracking only. $\pi_1$: tracking and foot-contact reward. Convergence (conv.) is defined as the number of environment steps until the success rate (SR) first reaches $50\%$ on a sliding evaluation window. We display the mean standard deviation given over 5 seeds. Best improvement in \textbf{bold}.}

        \resizebox{\linewidth}{!}{%
            \setlength{\aboverulesep}{1pt}
            \setlength{\belowrulesep}{2pt}
            \begin{tabular}{lcccc}
                \toprule
                \multirow{2}{*}{Motion}
                & \multicolumn{2}{c}{Max SR (\%) $\uparrow$}
                & \multicolumn{2}{c}{Conv. ($10^8$ steps) $\downarrow$} \\
                \cmidrule(lr){2-3} \cmidrule(lr){4-5}
                & $\pi_{0}$ & $\pi_{1}$ & $\pi_{0}$ & $\pi_{1}$ \\
                \midrule
                Flat    & $99.0_{\scriptscriptstyle \pm 0.1}$ & $99.0_{\scriptscriptstyle \pm 0.1}$ & $0.90_{\scriptscriptstyle \pm 0.01}$ & $0.84_{\scriptscriptstyle \pm 0.01}$ \\
                Box up  & $89.0_{\scriptscriptstyle \pm 0.1}$ & $90.0_{\scriptscriptstyle \pm 0.2}$ & $1.23_{\scriptscriptstyle \pm 0.07}$ & $1.08_{\scriptscriptstyle \pm 0.02}$ \\
                Beam    & $96.0_{\scriptscriptstyle \pm 0.1}$ & $96.0_{\scriptscriptstyle \pm 0.1}$ & $0.82_{\scriptscriptstyle \pm 0.02}$ & $0.75_{\scriptscriptstyle \pm 0.02}$ \\
                S.Stones & $67.0_{\scriptscriptstyle \pm 1.0}$ & $\boldsymbol{72.0}_{\scriptscriptstyle \pm 1.0}$ & $7.14_{\scriptscriptstyle \pm 0.65}$ & $\boldsymbol{5.72}_{\scriptscriptstyle \pm 0.21}$ \\
                \bottomrule
            \end{tabular}%
        }

        \label{tab:footlock_ablation}
    \end{minipage}
    \hfill%
    \begin{minipage}[t]{0.48\textwidth}
        \centering
        \vspace{-0pt}

        \captionof{figure}{Reference motion precision ablation. Different levels of Gaussian noise applied to root position of reference motion. Training performance is highly reduced above $\sigma =\SI{0.05}{m}$ and impossible at errors above $\SI{0.1}{m}$ (evaluated over 5 seeds each).}
        \label{fig:precision}

        \includegraphics[
            width=\linewidth,
            height=0.25\textheight,
            keepaspectratio
        ]{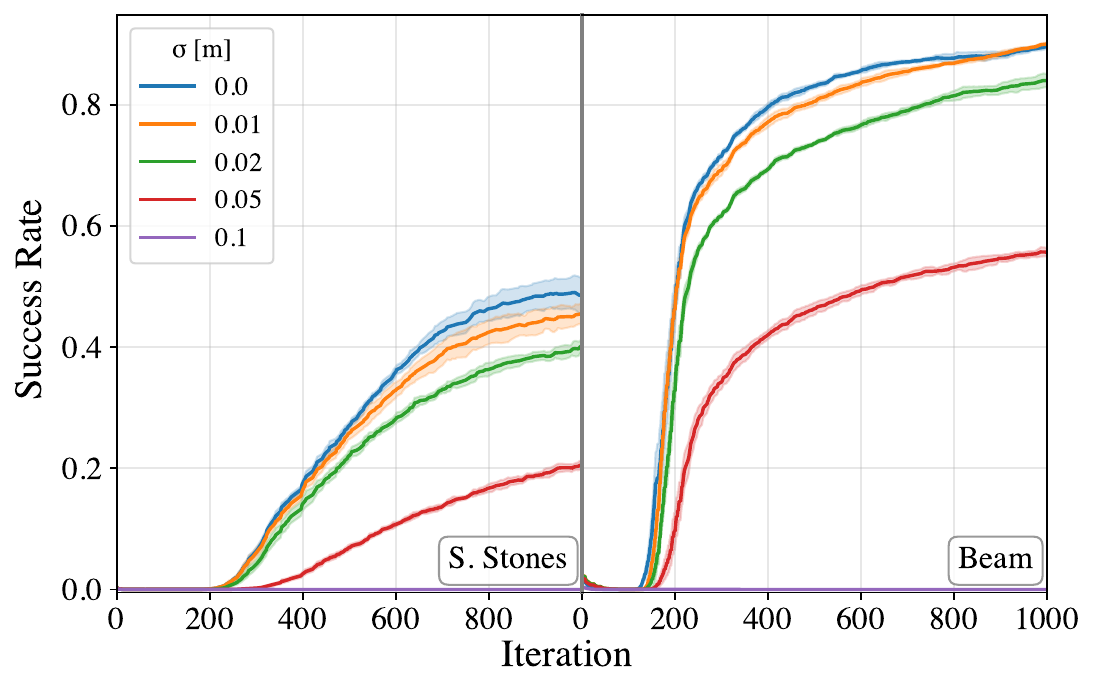}
    \end{minipage}%
    \vspace{-10pt}
\end{table*}

\begin{table}[htbp]
    \centering 
    \vspace{-15pt}
    \captionsetup{labelfont=bf, font=small, skip=2pt}
    \caption{The stages and timeline of the Real2Sim2Real pipeline with its specifications and failure cases. The listed times illustrate the workflow for a standard 20-minute motion capture and processing session.}

    \definecolor{stage1}{HTML}{E1BEE7} 
    \definecolor{stage2}{HTML}{C5D9F1} 
    \definecolor{stage3}{HTML}{B2DFDB} 
    \definecolor{stage4}{HTML}{C8E6C9} 
    \definecolor{stage5}{HTML}{FFE082} 
    \definecolor{stage6}{HTML}{FFCCBC} 

    \noindent\resizebox{\linewidth}{!}{
    \begin{tikzpicture}[font=\scriptsize]
    
    \def\barw{9}
    \def\barh{0.22} 
    \def\gap{0.05}  
    
    \pgfmathsetmacro{\tA}{120} 
    \pgfmathsetmacro{\tB}{90} 
    \pgfmathsetmacro{\tC}{20}  
    \pgfmathsetmacro{\tD}{20}  
    \pgfmathsetmacro{\tE}{720} 
    \pgfmathsetmacro{\tF}{180} 
    
    \pgfmathsetmacro{\sSum}{\tA+\tB+\tC+\tD+\tE+\tF}
    
    \pgfmathsetmacro{\wA}{\barw*\tA/\sSum}
    \pgfmathsetmacro{\wB}{\barw*\tB/\sSum}
    \pgfmathsetmacro{\wC}{\barw*\tC/\sSum}
    \pgfmathsetmacro{\wD}{\barw*\tD/\sSum}
    \pgfmathsetmacro{\wE}{\barw*\tE/\sSum}
    \pgfmathsetmacro{\wF}{\barw*\tF/\sSum}
    
    \pgfmathsetmacro{\xA}{0}
    \pgfmathsetmacro{\xB}{\xA+\wA}
    \pgfmathsetmacro{\xC}{\xB+\wB}
    \pgfmathsetmacro{\xD}{\xC+\wC}
    \pgfmathsetmacro{\xE}{\xD+\wD}
    \pgfmathsetmacro{\xF}{\xE+\wE}
    
    \tikzset{seg/.style={rounded corners=1pt, draw=black!20, line width=0.3pt}}
    
    \draw[draw=none] (0,-0.12) rectangle (\barw,\barh);
    
    \fill[seg, fill=stage1] ({\xA+\gap/2},0) rectangle ({\xB-\gap/2},\barh);
    \fill[seg, fill=stage2] ({\xB+\gap/2},0) rectangle ({\xC-\gap/2},\barh);
    \fill[seg, fill=stage3] ({\xC+\gap/2},0) rectangle ({\xD-\gap/2},\barh);
    \fill[seg, fill=stage4] ({\xD+\gap/2},0) rectangle ({\xE-\gap/2},\barh);
    \fill[seg, fill=stage5] ({\xE+\gap/2},0) rectangle ({\xF-\gap/2},\barh);
    \fill[seg, fill=stage6] ({\xF+\gap/2},0) rectangle ({\barw-\gap/2},\barh);
    
    \node[black] at ({(\xA+\xB)/2},{\barh/2}) {\textbf{2h}};
    \node[black] at ({(\xB+\xC)/2},{\barh/2}) {\textbf{1.5h}};
    \node[black] at ({(\xC+\xD)/2},{\barh/2}) {}; 
    \node[black] at ({(\xD+\xE)/2},{\barh/2}) {}; 
    \node[black] at ({(\xE+\xF)/2},{\barh/2}) {\textbf{4-16h}};
    \node[black] at ({(\xF+\barw)/2},{\barh/2}) {\textbf{2-3h}};
    
    \draw[->, >=stealth, draw=black!100, line width=0.6pt] (0.05, -0.06) -- (\barw-0.05, -0.06);
    
    \end{tikzpicture}%
    }
    
    \vspace{-1pt} 
    
    \noindent\resizebox{\linewidth}{!}{%
    \small
    \setlength{\tabcolsep}{3pt} 
    \renewcommand{\arraystretch}{1.1} 
    
    \setlength{\aboverulesep}{0pt}
    \setlength{\belowrulesep}{0pt}
    
    \begin{tabular}{l l l}
    \textbf{Stage} & \textbf{Specs} & \textbf{Failure Cases}\\
    \midrule
    \cellcolor{stage1}\textbf{Capture}     & max. 0.5h per seq. (hard limit by IMU suit) & long-term IMU drift; WiFi connection\\
    \cellcolor{stage2}\textbf{PProcess} & manual scene postprocessing (20min); ARIA MPS (1.5h) & Acc. threshold of Aria; featureless env\\
    \cellcolor{stage3}\textbf{Recon}  & auto spatial and temporal alignment & low init. img overlay; init. sound disturb. \\
    \cellcolor{stage4}\textbf{Retarget}    & GMR \cite{ze2025gmr}; auto scene crop and align (alt. OmniRetarget \citep{yang2025omniretarget}) & human2robot gap (high objects, ...)\\
    \cellcolor{stage5}\textbf{Train}        & Perceptive BeyondMimic Tracking \citep{liao2025beyondmimicmotiontrackingversatile} (RTX4090 24GB) & high robot-scene penetration or FFR\\
    \cellcolor{stage6}\textbf{Deploy}      & Hardware: G1 EDU Standard & Sim2Real height scan; init. mismatch \\
    \bottomrule
    \end{tabular}%
    }

    \label{tab:pipeline_timeline}
    \vspace{-6pt}
\end{table}

\subsection{Human Mesh Recovery}
\begin{figure}[htbp]
    \centering
    \vspace{-10pt}
    \captionsetup{labelfont=bf, font=small, skip=3pt}
    
    \captionof{table}{Benchmarking of mesh recovery methods on our dataset across different modalities. We evaluate success rate (SR), variants of MPJPE in mm (procrustes-aligned [PA-M.], world [W-M.], and world-aligned [WA-M.]), root translation error (RTE), chamfer distance (CD), precision (Prec), foot sliding (FS) in mm, foot floating rate (FFR) and jitter in $m/s^3$.}
    \label{tab:combined_hmr_benchmarks}
    
    \scriptsize 
    \setlength{\tabcolsep}{4pt} 
    \renewcommand{\arraystretch}{1.2} 
    \resizebox{\columnwidth}{!}{%
    \setlength{\aboverulesep}{0pt}
    \setlength{\belowrulesep}{0pt}
    \begin{tabular}{ll c cc ccc cc ccc}
        \toprule
        & & & \multicolumn{2}{c}{\textbf{Local}} & \multicolumn{3}{c}{\textbf{Global}} & \multicolumn{2}{c}{\textbf{Scene}} & \multicolumn{3}{c}{\textbf{Physics}} \\
        \cmidrule(lr){4-5} \cmidrule(lr){6-8} \cmidrule(lr){9-10} \cmidrule(lr){11-13}
        \textbf{Mod.} & \textbf{Methods} & SR(\%) $\uparrow$ & MPJPE $\downarrow$ & PA-M. $\downarrow$ & W-M. $\downarrow$ & WA-M. $\downarrow$ & RTE (\%) $\downarrow$ & CD $\downarrow$ & Prec (\%) $\uparrow$ & FS $\downarrow$ & FFR (\%) $\downarrow$ & Jitt. $\downarrow$ \\
        \midrule
        
        \rowcolor{gray!10}
        & JOSH (ICLR'26)     & 47.72 & 104.0 & 67.3 & 356.0 & 149.1 & 2.20 & 1.40 & 66.87 & 25.99 & 20.91 & 2.38 \\
        \rowcolor{gray!10}
        & Human3R (ICLR'26)  & 87.06 & 138.6 & 63.2 & 641.4 & 289.2 & 5.08 & 3.28 & 14.12 & 74.21 & 89.62 & 10.71 \\
        \rowcolor{gray!10}
        \multirow{-3}{*}{Exo} & VideoMimic (CoRL'25) & 85.08 & 112.7 & 62.6 & 381.1 & 190.5 & 1.50 & 0.66 & 46.83 & 23.13 & 85.06 & 1.56 \\
        Ego & EgoAllo (CVPR'25)  & 100 & 160.1 & 110.9 & 386.5 & 165.8 & 0.24 & -- & -- & -- & -- & 0.14 \\
        
        
        \bottomrule
    \end{tabular}
    }

    \vspace{3pt} 

    \includegraphics[trim={0cm 0cm 0cm 0cm}, clip, width=\linewidth]{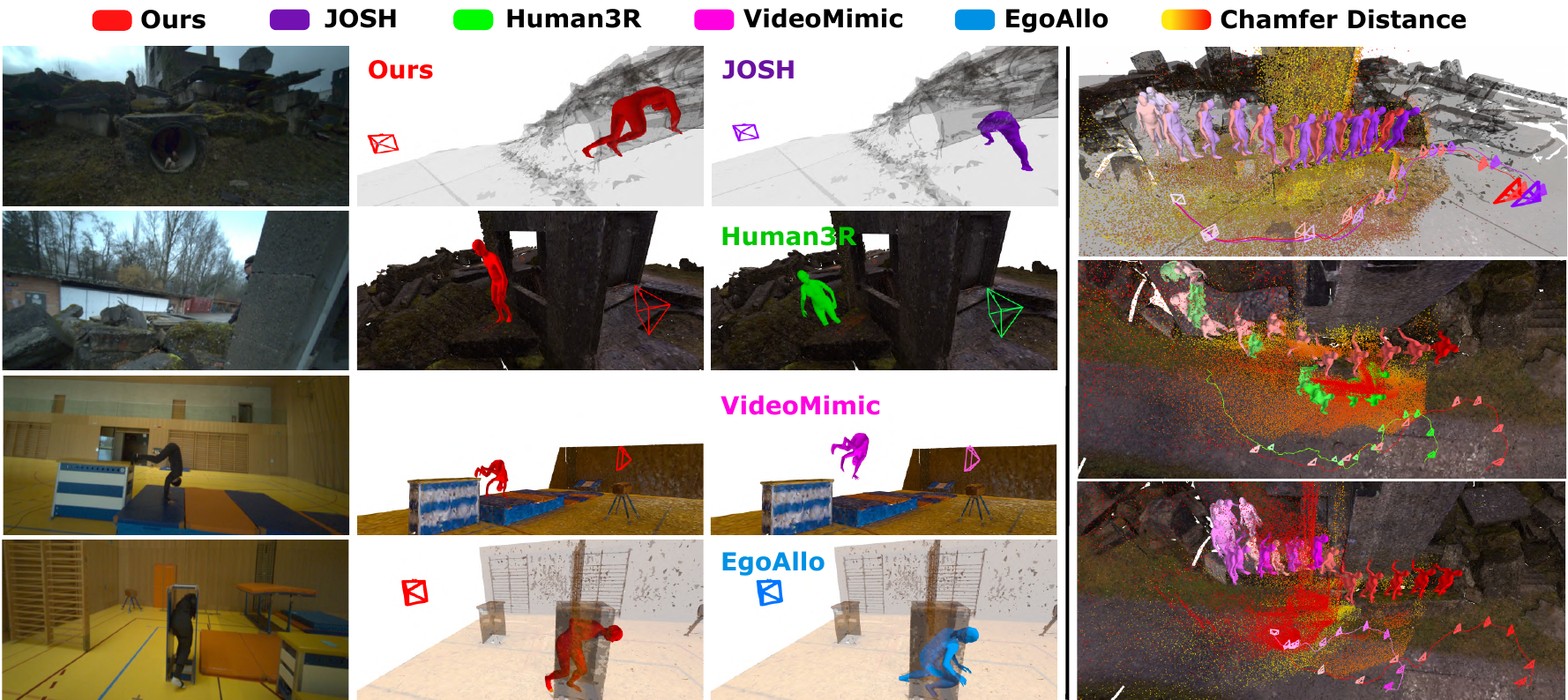}
    \captionsetup{labelfont=bf, font=small, skip=3pt, justification=justified}
    \captionof{figure}{Failure cases of SOTA human mesh recovery methods under challenging environmental conditions present in our dataset. From left to right: Observer Aria image frames. EgoHTR reconstructions. Failure cases of SOTA methods induced by complex scene interactions. Visualizing the global motion estimates of the exocentric methods \citep{liu2026joint, chen2025human3r, videomimic} (top to bottom order), together with a color-encoded chamfer distance of the reconstructed scene point cloud.}
    \label{fig:quali_mesh_recovery}
    \vspace{-15pt}
\end{figure}

The multimodality of our dataset enables comprehensive benchmarking of 4D human-scene reconstruction across exocentric, egocentric, and inertial paradigms and allows a direct comparison of the methods, see \cref{tab:combined_hmr_benchmarks}. 
Current exocentric monocular approaches struggle with traditional in-the-wild challenges. The low success rate of exocentric methods \citep{liu2026joint, chen2025human3r, videomimic} expose severe vulnerabilities to occlusions (shown in \cref{fig:quali_mesh_recovery}, left) and motion blur. Because our dataset includes such challenging cases, EgoHTR uniquely facilitates qualitative and quantitative evaluation under these harsh conditions. The global and scene metrics track the global drift important for the long-term consistency of the method (visualized in \cref{fig:quali_mesh_recovery}, right). 
The physics metrics evaluate the plausibility and temporal smoothness of the reconstructed motion. 
Given the comparison in \cref{tab:combined_hmr_benchmarks}, especially exocentric end-to-end models \citep{chen2025human3r, sur2026unicon3rcontactaware3dhumanscene} lack an inherent understanding of human-scene interactions, while optimization-based pipelines struggle to generalize under harsh conditions. 
The benchmarking on other modalities reveals additional limitations. 
SOTA egocentric models \cite{li2023ego, yi2025egoallo, jiang2024egoposer, hollidt2024egosim}, trained primarily on flat ground motions, fail to capture dynamic motions in complex environments, as reflected by their high local joint errors.
Moreover, while IMU-based posers \cite{PIPCVPR2022, jiang2022transformer, hollidt2026ultra} bypass visual occlusions, temporal drift causes them to suffer from physically impossible scene interactions, crucial for scene-aware motion reconstruction \citep{hsu2026imu4d}.\newline
To overcome this, our dataset can serve as a resource to fine-tune these networks and potentially avoid expensive post-hoc optimization \cite{liu2026joint, videomimic}. A promising direction we leave for future work. This idea could be further extended to motion synthesis, see \cref{app:human_motion_generation}. 

\section{Limitations} \label{sec:limitations}

\textbf{Dataset Statistics  }
Our current dataset serves as an effective benchmark and fine-tuning dataset, but lacks the scale and diversity for comprehensive large-scale training. 
However, our pipeline is designed to be highly extensible, allowing for multi-person capture, text annotations via audio streams, and continuous integration of community-sourced sequences.

\textbf{Human-Scene Reconstruction  }
Currently, our data pipeline is restricted to static environments devoid of articulated objects.
While the egocentric glasses allow hand tracking, it is not incorporated in the body model. 
Further, we do not perform post-capture joint human-scene optimization.  
However, our framework provides the interface to incorporate an extended joint optimization \cite{PROX2019, Guzov2021HPS} to improve interaction fidelity and correct for relative IMU drift. 
Lastly, the pipeline performance is subject to the physical constraints of the hardware, with potential failures in featureless environments or during high-acceleration maneuvers that challenge sensor localization capabilities. 

\textbf{Perceptive Whole-Body Control} 
While our pipeline successfully enables Real2Sim2Real transfer in complex terrain, training separate expert policies for distinct tasks prevents generalized motion control. Further, retargeting highly dynamic behaviors (e.g. steep climbs or long jumps) remains challenging due to the human-to-G1 embodiment gaps. Additional pipeline specifications and limitations are outlined in \cref{tab:pipeline_timeline}.

\section{Conclusion}
\label{sec:conclusion}
We introduced EgoHTR, to the best of our knowledge, the first dataset providing 4D human-scene motion on rough terrain under in-the-wild conditions. 
With a multi-sensor setup, we can reconstruct accurate SMPL body model sequences alongside high-resolution multi-modal contextual data. 
EgoHTR addresses the critical data gap of grounded, versatile demonstrations within an unconstrained environmental context. 
The dataset not only highlights the immediate values as a resource for learning context-aware robotic locomotion and motion synthesis, but also establishes a challenging benchmark for state-of-the-art human mesh recovery.
By open-sourcing our capture and reconstruction pipeline, we aim to build up a dataset capable of training context-aware foundation models for robot learning.
In future work, we plan to extend our data acquisition pipeline to encompass human-object interactions, integrating fine-grained hand and object tracking and leveraging the dataset's multi-modality to advance toward 4D world reconstruction for humanoid loco-manipulation.

\paragraph{Acknowledgments}
We thank Ewa Zurmühle, Daniela Brandes, Ruth Brandes, Xinyi Chen, Haig Conti Georges Sajelian, Simon Balmer, Mikail Gedik, and Manthan Patel for their support during the data capture process. 
We are also grateful to the Mobile Robotics Lab for compute and IT infrastructure and to the ETH AI Center for providing the Aria glasses and the Rokoko suit. 
This work was supported by European Union’s Horizon Europe research and innovation program under grant agreement number 101214398 (ELLIOT), fully funded by the Swiss State Secretariat for Education, Research and Innovation (SERI).

\bibliography{bibliography/references}

\clearpage 
\appendix
\crefalias{section}{appendix} %
\crefalias{subsection}{appendix} %
\counterwithin{figure}{section}
\counterwithin{table}{section}
\counterwithin{equation}{section}
\phantomsection

\begin{center}
    {\large \bf EgoHTR: Egocentric 4D Demonstrations of Human Terrain Traversal} \\[1.0ex]
    {\large Supplementary Material} \par
    \vspace{1.0em}
\end{center}
\addtocontents{toc}{\protect\setcounter{tocdepth}{2}}

This supplementary document provides additional technical details, extended evaluations, and qualitative results to support the main manuscript. Specifically, Section A offers an extended review of related work. Section B expands upon our methodology, providing deeper insights into the hardware setup, the reconstruction pipeline, and ground-truth motion capture acquisition. Section C presents a comprehensive overview and extended statistics of the EgoHTR dataset. Finally, Section D details our extended experiments and elaborates on downstream applications.

\section{Extended Related Work} \label{app:extended_related_work}
\paragraph{Human Motion Dataset}
\Cref{tab:ext_dataset_comparison_final} provides an overview of the most recent human motion datasets. 
Early laboratory marker-based systems \citep{AMASS:ICCV:2019, harvey2020lafan1} and text-conditioned motion databases provide high-fidelity kinematics but either lack environmental context entirely \citep{Guo2022HML3D} or fail to make it publicly available \citep{Black_CVPR_2023, bones2026aidatasets}. 
To bridge this gap, exocentric RGB-D approaches incorporate scene interactions through both monocular \citep{zhang2025motion, lin2023motionx, Zhu_2023_ICCV, PROX2019} and multi-view \citep{zhang2022egobody, xu2023inter, grauman2024egoexo4dunderstandingskilledhuman, kim2024parahome, li2023object, RICHHuang:CVPR:2022, khirodkar2023egohumans, bones2026aidatasets} setups, enabling the study of whole-body, human-human, and human-object interactions. However, these systems remain spatially constrained to indoor, or structured locations. 
To enable scaling to unbounded capture volumes, researchers have increasingly adopted wearable and egocentric in-the-wild motion capture. 
While inertial and monocular fusion \citep{vonMarcard2018, kaufmann2023emdb, yang2024divatrackdiversebodiesmotions} enables location independent camera-relative and global human motion estimation, the reconstruction remains ungrounded to surrounding scene context. 
Other methodologies address this problem by fusing egocentric capture with pre-scanned environments \citep{Guzov2021HPS, cong2024laserhuman}, leveraging multi-modal sensor devices \citep{zheng2022gimo, ma2024nymeriamassivecollectionmultimodal, detone2026nymeriaplusenrichingnymeriadataset} or incorporating LiDAR tracking \citep{Dai_2023_CVPR} to capture long-range activities in the scene. 
Although datasets like \citep{Guzov2021HPS, zheng2022gimo, RICHHuang:CVPR:2022, Dai_2023_CVPR} enable 4D human-scene reconstruction, they suffer from missing sensor modalities -- lacking exocentric or egocentric vision — and environmental biases, being restricted to structured or spatially confined spaces. 
Consequently, these bottlenecks restrict future multi-modal research in human motion reconstruction and humanoid control for in-the-wild terrain, a gap that our proposed dataset is specifically designed to target.

\paragraph{Data-Driven Human Mesh Recovery}
The evolution of human mesh recovery (HMR) has been fundamentally shaped by the availability of diverse training data. 
In human-only reconstruction, progress has been driven by datasets supporting varied modalities, including single-view images \citep{sarandi2024neural, multi-hmr2024, patel2025camerahmr, wang2025prompthmrpromptablehumanmesh}, temporal exocentric \citep{wham:cvpr:2024, wang2024tram, shen2024gvhmr, li2025genmo, yang2026sam}, egocentric video sequences \citep{li2023ego, yi2025egoallo} as well as sparse inertial measurement units (IMUs) \citep{PIPCVPR2022, yi2025improving}. 
Despite these successes in reconstructing the human figure in isolation, research is increasingly shifting toward 4D human-scene reconstruction. 
Many recent methods process human motion and scene geometry through independent modules, necessitating complex, post-hoc optimization to align the reconstructed human with the 3D environment \citep{videomimic, zhang2026meshmimicgeometryawarehumanoidmotion, wang2026embodmocapinthewild4dhumanscene, liu2026joint}. 
Conversely, approaches that attempt to implicitly learn the 4D reconstruction, such as unified vision models \citep{chen2025human3r, sur2026unicon3rcontactaware3dhumanscene} and terrain-aware IMU posers \citep{jiang2022transformer}, suffer from the human-scene data gap. 
They struggle to capture physically plausible real-world interactions. 
Consequently, there remains a critical need for multi-modal datasets that capture in-the-wild human-scene motion sequences with high-resolution contextual data.

\begin{table}[htbp]
    \centering
    \captionsetup{labelfont=bf, font=small, skip=5pt, justification=justified}
    \caption{Extended Version of Human Motion Dataset comparison.}
    \label{tab:ext_dataset_comparison_final}
    
    \begin{minipage}[c]{0.80\linewidth}
        \scriptsize
        \setlength{\tabcolsep}{3pt} 
        \renewcommand{\arraystretch}{0.9} 
        
        \resizebox{\linewidth}{!}{
        \begin{tabular}{l l c c c c c c c c c c c c}
        \toprule
            \textbf{Dataset} & \textbf{Year} & \multicolumn{3}{c}{\textbf{Scale}} & \multicolumn{6}{c}{\textbf{Modalities}} & \multicolumn{3}{c}{\textbf{Location}} \\
        \cmidrule(lr){3-5} \cmidrule(lr){6-11} \cmidrule(lr){12-14}
            & & \faClock & \faImages & \faHourglassHalf & 
            \faMale & \faGlasses &  \faBinoculars &  \faVideo & \faEye & \faTree & 
            \faHome & \faCloudSun & \faMountain \\
        \midrule
            \rowcolor{Gray}
            3DPW \citep{vonMarcard2018} & 2018 & 0.5 & 0.05 & 0.48 & \cm & \cm & \cm & & & & & \cm & \\
            AMASS \citep{AMASS:ICCV:2019} & 2019 & 42 & 0.9 & 0.22 & \cm & & & & & & \cm & & \\
            \rowcolor{Gray}
            PROX \citep{PROX2019} & 2019 & 0.92 & 0.1 & 4.6 & \cm & & & \cm & & \cm & \cm & & \\  
            LaFAN1 \citep{harvey2020lafan1} & 2020 & 4.6 & 0.5 & 3.6 & \cm & & & \cm & & & \cm & & \\
            \rowcolor{Gray}
            HPS \citep{Guzov2021HPS} & 2021 & 4.5 & 0.3 & 8.2 & \cm & \cm & & & & \cm & \cm & \cm & \\
            EgoBody \citep{zhang2022egobody} & 2022 & 2.03 & 0.22 & 1 & \cm & \cm & & \cm & \cm & \cm & \cm & & \\
            \rowcolor{Gray}
            HML3D \citep{Guo2022HML3D} & 2022 & 28.6 & 2.06 & 0.12 & \cm & & & & & & \cm & & \\
            RICH \citep{RICHHuang:CVPR:2022} & 2022 & 5.34 & 0.58 & 2.26 & \cm & & & \cm & & \cm & \cm & \cm \\
            \rowcolor{Gray}
            GIMO \citep{zheng2022gimo} & 2022 & 1.2 & 0.125 & 0.32 & \cm & \cm & & & \cm & \cm & \cm & & \\
            BEDLAM \citep{Black_CVPR_2023} & 2023 & 14.8 & 1.6 & 0.38 & \cm & & & & & \cm & \cm & \cm \\
            \rowcolor{Gray}
            OMOMO \citep{li2023object} & 2023 & 10 & 4.32 & 40.4 & \cm & & & \cm & & \cm & \cm & & \\
            EgoHuman \citep{khirodkar2023egohumans} & 2023 & 1.73 & 0.41 & 0.5 & \cm & \cm & & \cm & & $(\cm)$ & \cm & \cm & \\
            \rowcolor{Gray}
            EMDB \citep{kaufmann2023emdb} & 2023 & 0.97 & 0.11 & 0.72 & \cm & \cm & \cm & & & & \cm & \cm & \\
            SLOPER4D \citep{Dai_2023_CVPR} & 2023 & 1.38 & 0.1 & 5.5 & \cm & \cm & \cm & & & \cm & & \cm & \\
            \rowcolor{Gray}
            EgoExo4D \citep{grauman2024egoexo4dunderstandingskilledhuman} & 2024 & 1286 & 14 & 1.42 & & \cm & & \cm & \cm & $(\cm)$ & \cm & \cm & \cm \\
            ParaHome \citep{kim2024parahome} & 2024 & 8.1 & 1.7 & 2.34 & \cm & & & \cm & & \cm & \cm & & \\
            \rowcolor{Gray}
            Motion-X{\tiny ++} \citep{lin2023motionx, zhang2025motion} & 2025 & 180.9 & 19.5 & 0.11 & \cm & & \cm &  & & & \cm & \cm & \\
            NymeriaPlus \citep{detone2026nymeriaplusenrichingnymeriadataset} & 2026 & 300 & 260 & 15 & \cm & \cm & \cm & & \cm & $(\cm)$ & \cm & \cm & \\
            \rowcolor{Gray}
            BONES-SEED \citep{bones2026aidatasets} & 2026 & 288 & 124.5 & 0.12 & \cm & & & & & & \cm & & \\
            
            \midrule
            \textbf{EgoHTR (Ours)} & 2026 & 1.37 & 0.15 & 1.5 & \cm & \cm & \cm & \cm & \cm & \cm & \cm & \cm & \cm \\
        \bottomrule
        \end{tabular}
        }
    \end{minipage}
    \hfill
    \begin{minipage}[c]{0.18\linewidth}
        \scriptsize
        \renewcommand{\arraystretch}{1.0}
        \resizebox{\linewidth}{!}{
        \begin{tabular}{@{}c l@{}}
            \multicolumn{2}{@{}l}{\textbf{Scale}} \\
            \faClock & Hours \\
            \faImages & Frames (M) \\
            \faHourglassHalf & Avg. Seq. (min) \\
            \addlinespace
            \multicolumn{2}{@{}l}{\textbf{Modalities}} \\
            \faMale & Param. Body \\
            \faGlasses & Subject (ego) \\
            \faBinoculars & 2nd Person \\
            \faVideo & Fixed Cam. \\
            \faEye & Gaze \\
            \faTree & Scene Mesh\\
            \addlinespace
            \multicolumn{2}{@{}l}{\textbf{Location}} \\
            \faHome & Indoor \\
            \faCloudSun & Outdoor \\
            \faMountain & Rough Terrain \\
            \addlinespace
            \multicolumn{2}{@{}p{0.9\linewidth}@{}}{$(\cm)$ limited scene reconstruction} \\
        \end{tabular}
        }
    \end{minipage}
\end{table}

\section{Methodology} \label{app:sec_methodlogy}

\subsection{Hardware} \label{app:sec_method}

\begin{figure}[htbp]
    \centering
    \captionsetup{labelfont=bf, font=small, skip=3pt, justification=justified}
    \caption{Hardware used for the capturing of the EgoHTR Dataset: Egocentric motion capture suit (Smartsuit Pro II) (1), marker-based motion capture suit for the pseudo ground truth data capture with a Qualisys motion capture system (2), 3D scanner (BLK2GO) (3) and the Aria glasses Gen. 1 (subject (4) and object (5)).}
    \includegraphics[trim={6cm 8cm 5.5cm 6cm}, clip, width=\textwidth]{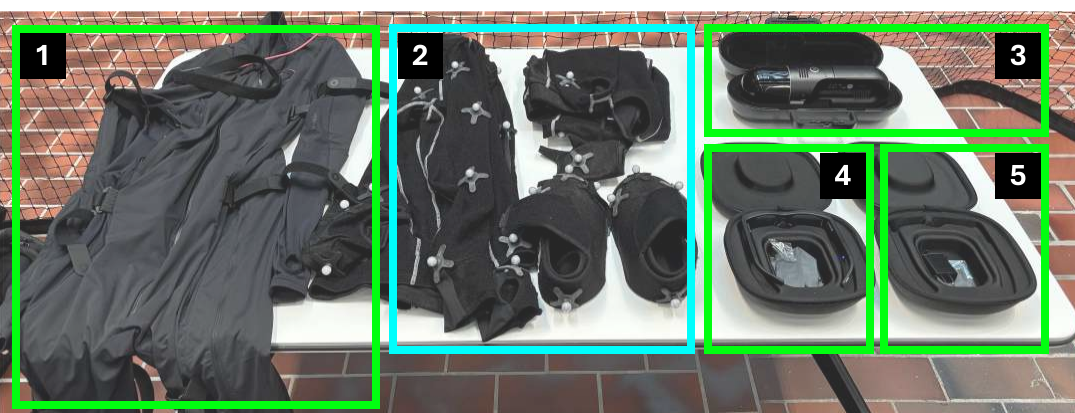}
    \label{fig:hardware}
    \vspace{-10pt}
\end{figure}

\begin{table}[h]
    \centering
    \captionsetup{labelfont=bf, font=small, skip=5pt, justification=justified}
    \caption{Hardware configurations and corresponding data streams provided in the dataset.}
    \label{tab:hardware_data_streams}
    
    \scriptsize 
    \setlength{\tabcolsep}{8pt} 
    \renewcommand{\arraystretch}{1.1} 
    
    \begin{tabular}{@{} l l l l @{}}
    \toprule
        \textbf{Sensor / Hardware} & \textbf{Data Stream / Modality} & \textbf{Format} & \textbf{Spec / Freq.} \\
    \midrule
        
        \textbf{Project Aria Glasses} & RGB Frames & VRS & 1408$\times$1408, \SI{30}{fps} \\
        (Subject \& Observer) & SLAM Trajectory ($\mathbf{C}_A$) & MPS & \SI{1000}{fps} \\
        & Semi-dense Point Cloud ($\mathcal{P}_A$) & MPS & - \\
        & Hand Tracking & MPS & - \\
        & Audio Stream & VRS/MP4 & - \\
        
    \midrule
        \rowcolor{Gray}
        \textbf{Ego. MoCap Suit} & Relative Joint Kinematics ($\mathbf{q}_E$, $\mathbf{p}_E$) & BVH & \SI{30}{fps} \\
        \rowcolor{Gray}
        (Rokoko) & Raw IMU Measurements & CSV & \SI{100}{fps} \\
        
    \midrule
        
        \textbf{3D Scene Scanner} & Dense Point Cloud & E57 & High-res \\
        (Leica BLK2GO) & Colorized 3D Mesh & OBJ, MTL, PNG & High-fidelity \\
        & Initial Pose Image ($\mathbf{I}_{C_{S,0}}$) & PNG & High-res \\
        
    \midrule
        \rowcolor{Gray}
        \textbf{Fixed Exocentric Cam.} & 3rd-Person RGB Video & MP4 & 3840$\times$2160 \\
        \rowcolor{Gray}
        (GoPro Hero 10) & & & \\
        
    \midrule
        
        \textbf{Exocentric MoCap} & Marker-based Kinematics (Eval) & JSON & 300 fps \\
        
    \bottomrule
    \end{tabular}
\end{table}

\paragraph{Project Aria Glasses} Both Aria devices (worn by the subject $A_{sub}$ and, optionally, the observer $A_{obs}$) use recording profile 15. This configuration records 1408$\times$1408 RGB frames at \SI{30}{fps} and provides a closed-loop SLAM trajectory of the glasses camera poses $\mathbf{C}_A$ at \SI{1000}{fps}, together with a processed semi-dense point cloud $\mathcal{P}_A$. Data post-processing is performed using the official SLAM and Multi-SLAM Machine Perception Services (MPS) provided by Meta.

\paragraph{Egocentric Motion Capture Suit} The egocentric IMU-based motion capture suit (Rokoko) outputs relative joint angles and offsets in a BVH format at \SI{30}{fps}, while also logging raw IMU measurements at \SI{100}{fps}. Using forward kinematics, we compute per-frame global joint quaternions $\mathbf{q}_E$ and 3D joint positions $\mathbf{p}_E$. To reduce drift during recording, the suit is recalibrated in a standard I-pose at the start of each sequence. We additionally record subject-specific anthropometric measurements to configure the corresponding digital avatars (see \cref{tab:subject_measurements_id_sorted}).

\paragraph{3D Scanner} For high-resolution scene geometry, we use a Leica BLK2GO handheld imaging laser scanner. The device captures dense 3D point clouds, alongside continuous high-resolution imagery and a 6-DoF trajectory of the scanner. Using the official Leica Cyclone REGISTER 360 software, the raw SLAM output is post-processed into a high-fidelity, colorized 3D mesh. The initial scanner camera pose $\mathbf{C}_{S,0}$ and its associated image $\mathbf{I}_{C_{S,0}}$ provide the geometric anchor used for the global alignment of the egocentric sensors. The ground-projected start position of the scanner defines the global origin, establishing the unified world frame $\mathcal{W}$ to which all modalities are aligned.

\paragraph{Fixed Camera} To provide an additional exocentric view of the human-scene interaction, we deploy a fixed-position GoPro Hero 10 camera. The static camera records high-resolution video of the environment and offers a stable third-person perspective.  

\subsection{Body model}
We parameterize the human subject using the Skinned Multi-Person Linear eXpressive (SMPL-X) model, a differentiable framework that jointly represents body, face, and hands \citep{SMPL-X:2019}. 
SMPL-X outputs a dense body mesh via the mapping function $M(\boldsymbol{\theta}, \boldsymbol{\beta}, \boldsymbol{\psi}, \boldsymbol{t})$, where deformation is controlled by the kinematic pose parameters $\boldsymbol{\theta} \in \mathbb{R}^{165}$ (representing the axis-angle rotations of 55 joints), identity-dependent shape parameters $\boldsymbol{\beta} \in \mathbb{R}^{10}$, facial expression parameters $\boldsymbol{\psi} \in \mathbb{R}^{10}$, and a global translation $\mathbf{t} \in \mathbb{R}^3$  \citep{SMPL-X:2019}. \newline
In our setting, the motion capture suit provides pose information only for the 22 primary SMPL joints \citep{SMPL:2015}. 
We therefore fix all remaining hand and facial parameters to the identity pose, providing a modular baseline for future extensions. 
Since body shapes ($\boldsymbol{\beta}$) and facial expression ($\boldsymbol{\psi}$) are assumed constant within a sequence, we estimate only the articulated pose $\boldsymbol{\theta}$ and the global translation $\boldsymbol{t}$. Shape parameters are obtained using state-of-the-art toolkits \citep{Black_CVPR_2023, SMPL-X:2019}.

The body model retargeting (\cref{fig:method}, Ia) converts global joint orientations from the egocentric source skeleton, represented as quaternions $\mathbf{q}_{E}$, into the relative rotation matrices required by the target SMPL-X, $\boldsymbol{\theta} = f(\mathcal{R}(\mathbf{q}_{E,i}) ~\mathbf{C}_j)$, following the work of \citet{ze2025gmr}. 
To compensate for systematic axis misalignment between the two skeleton, we introduce a per-joint offset matrix $\mathbf{C}_j \in SO(3)$. 
Let $j$ denote a target joint and $i$ be its corresponding source joint under the joint mapping $\mathcal{M}: i \mapsto j$, whereas $\mathcal{R}: \mathbb{H} \to SO(3)$.

To mitigate for joint position errors introduced during retargeting, we refine the pose using an optimization-based inverse kinematics (IK) step (\cref{fig:method}, Ib). 
Specifically, we optimize the SMPL-X pose parameters $\boldsymbol{\theta}$ such that the model joint positions match target joint positions $\mathbf{p}_{E}$, obtained via forward kinematics from the motion capture suit's relative joint angles. 
Due to structural differences between the source and target skeleton, the loss is restricted to a subset  $\mathcal{L}$ of limb joints:

\begin{equation} \label{equ:ik}
    \boldsymbol{\theta}^* = \arg\min_{\boldsymbol{\theta}} \sum_{j \in \mathcal{L}} \left\| \mathcal{J}(\boldsymbol{\theta)}_j - \mathbf{p}_{E_j} \right\|^2
\end{equation}

where $\mathcal{J}(\boldsymbol{\theta})_j$ denotes the 3D position of the $j$-th SMPL-X joint and $ \mathbf{p}_{E_j}$ the corresponding target joint position.\\

\subsection{Temporal Alignment} The clapping motion generates near-simultaneous, high-amplitude peaks in both the raw hand joint IMU signals (\SI{100}{Hz}) and the audio stream (\SI{48}{kHz}) recorded by the Aria glasses' microphone. The static camera (5) is aligned by manually identifying the clap frame using visual and acoustic cues.

\begin{figure}[htbp]
    \centering
    \captionsetup{labelfont=bf, font=small, skip=3pt, justification=justified}
    \caption{Temporal alignment analysis evaluating sensor timestamp drift between the Aria Glasses and the egocentric motion capture suit over a 260 s duration. Left: The time difference between sensors recorded over the course of the test. Right: The distribution of the expected time offsets for the left (LH) and right hand (RH) acceleration sensors.}
    \includegraphics[trim={0cm 6cm 0cm 7.5cm}, clip, width=\textwidth]{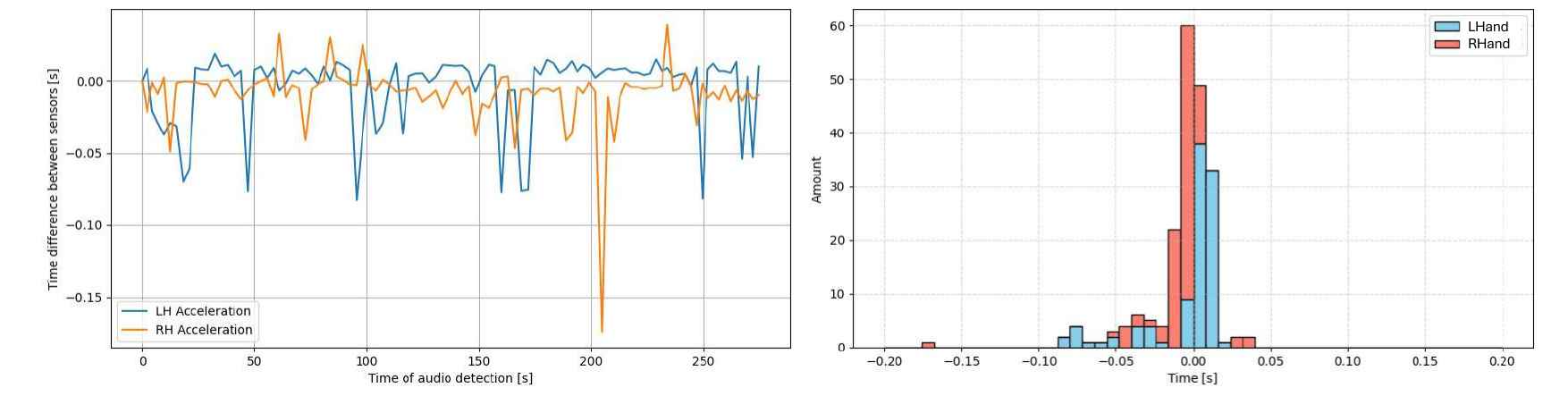}
    \label{fig:temporal_alignment}
\end{figure}

\subsection{Human2Robot Retargeting}
\label{sec:Human2Robot}
\begin{figure}[htbp]
    \centering
    \captionsetup{labelfont=bf, font=small, skip=3pt, justification=justified}
    \caption{Sequence of successive sampled frames of a retargeted motion from an SMPLX model to a G1 embodiment within robot-scaled environmental mesh. The stream illustrates retargeting of a backflip (A) and a handstand (B).}
    \includegraphics[trim={0cm 0cm 0cm 0cm}, clip, width=\textwidth]{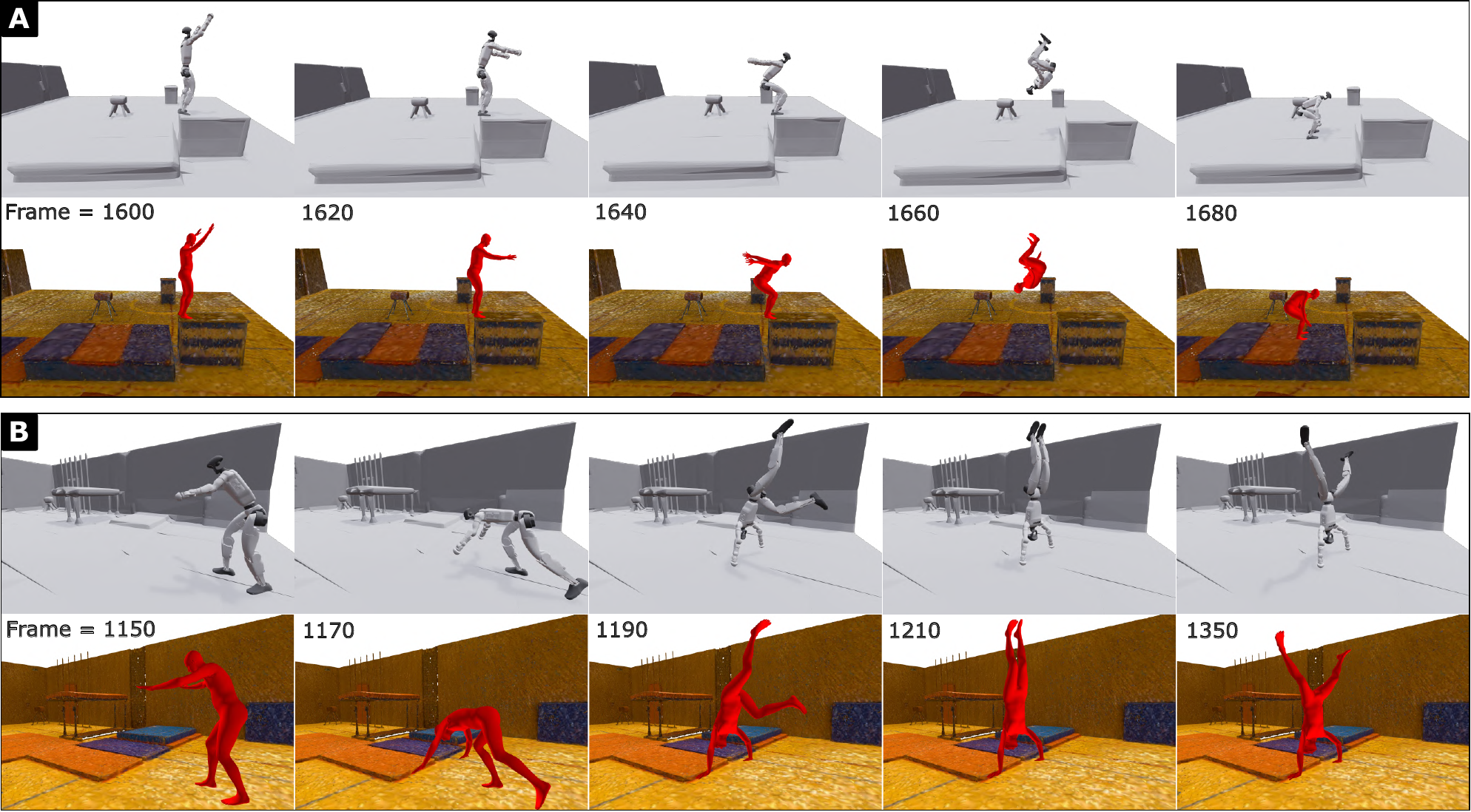}
    \label{fig:h2r}
\end{figure}

We used GMR \citep{ze2025gmr}, selected for its smooth outputs, to retarget the human motion to the robotic embodiment for the training of perceptive humanoid locomotion (described in \cref{app:training_pipeline}).\\
To further showcase and explore the dataset's broad usability, we also deployed OmniRetarget \citep{yang2025omniretarget} to successfully retarget the kinematic pose sequences within a convexified \citep{wei2022coacd} scene (examples in \cref{fig:h2r}).

\subsection{Motion Capture Ground Truth Acquisition} \label{sec:gt_acquisition}

We record ground truth motion in parallel to the human-scene reconstruction sequences using an optical motion capture system (Qualisys AB, Gothenburg, Sweden). The system tracks retroreflective markers with infrared cameras. The subject wears a dedicated suit with a predefined marker layout, from which the system reconstructs 3D marker trajectories with millimeter-level accuracy. The raw marker trajectories are processed with the proprietary Qualisys skeleton solver to obtain an initial kinematic skeleton. Finally, we retarget this skeleton to the SMPL-X body model and refine the skeleton mapping via a subsequent inverse kinematics (IK) optimization, following the same methodology detailed in \cref{sec:methodology}. 

\paragraph{Temporal Alignment}
To synchronize the exocentric motion capture stream with the egocentric sensor suite, we use the same distinct clapping event as for the motion capture suit. We estimate the hand joint acceleration by deriving the second derivative of the hand marker trajectories. As a cross-check, we additionally monitor a specific palm marker. The exact frame of the clap is confirmed by the momentary occlusion of this marker when the hands make contact.

\paragraph{Spatial Alignment}
We spatial align the marker-based and the egocentric sensor coordinates by tracking a dedicated marker mounted above the Aria glasses' camera. Over a fixed time window, we register this marker trajectory to the closed-loop Aria SLAM trajectory. For the coarse initialization, we align reconstructed skeleton's initial heading with the initial Aria camera heading both anchored at the initial Aria camera pose. 

We refine the coarse registration with Iterative Closest Point (ICP) $f_{\text{ICP}}$ on the temporally aligned trajectories from both systems, $\mathcal{T}_M$ and $\mathcal{T}_A$. Finally, we apply a constant offset transform $T_{\text{offset}}$ to account for the displacement between the tracked marker and the body model's root. This yields the final transform $T_{\mathcal{W}_A \mathcal{W}_M}$ mapping the exocentric motion capture frame $\mathcal{W}_M$ into the Aria world frame $\mathcal{W}_A$, as formalized in \cref{eq:mocap_align}:

\begin{equation} \label{eq:mocap_align}
T_{\mathcal{W}_A\mathcal{W}_M} = f_{\text{ICP}}(\mathcal{T}_A, \mathcal{T}_M) \cdot T_{\text{offset}}
\end{equation}



\clearpage
\section{Overview EgoHTR Dataset}
\begin{figure}[htbp]
    \centering
    \includegraphics[width=\textwidth]{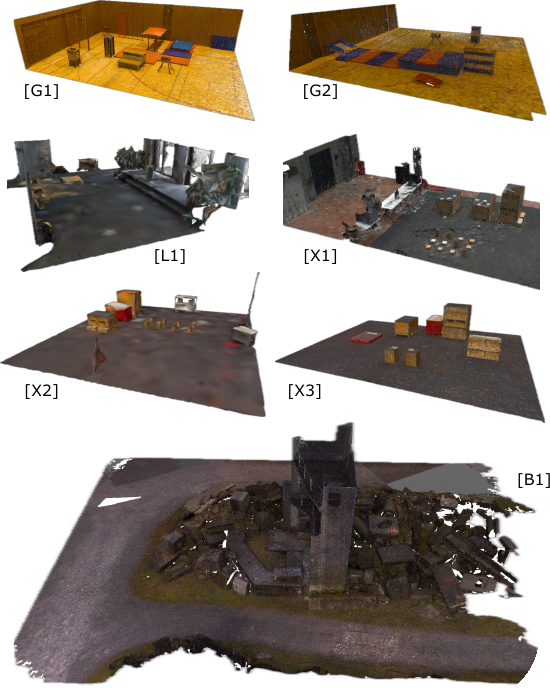}
    \captionsetup{labelfont=bf, font=small, skip=0pt, justification=justified}
    \vspace{6pt}
    \caption{High resolution scene meshes used in the EgoHTR dataset presenting different locations (G: Gym Hall, X: Robot Hall, B: BZB Debris field, L: LEE Office Room) and different setups (1, 2, 3)}
    \label{fig:scenes}
\end{figure}

\Cref{tab:dataset} provides a detailed overview of the EgoHTR dataset. \Cref{fig:scenes} illustrates all captured scene configurations, and \cref{tab:app_sequence_overview} summarizes the sequences considered for evaluation.

\begin{table}[htbp]
\centering
\captionsetup{labelfont=bf, font=small, skip=10pt, justification=justified}

    \begin{threeparttable}
    \caption{Overview of the EgoHTR Dataset, listing the features of the specific sequences in the corresponding scenes. The complete set can be split into a test subset used for quantitative evaluation or a multi-view subset. The latter consists of all sequences with an additional Aria glasses data stream}
    \label{tab:dataset}

    \scriptsize
    \setlength{\tabcolsep}{3.5pt} 
    \renewcommand{\arraystretch}{1.25}
    
    \begin{tabular}{l l c c c c c c c c c c c c c c c}
    \toprule
        & & & \multicolumn{2}{c}{\textbf{Subj.}} & \multicolumn{4}{c}{\textbf{Sequences}} & \multicolumn{6}{c}{\textbf{Sensors}} & \multicolumn{2}{c}{\textbf{Loc.}} \\
        \cmidrule(lr){4-5} \cmidrule(lr){6-9} \cmidrule(lr){10-15} \cmidrule(lr){16-17}
        \textbf{Scene} & \textbf{ID} & \textbf{Act.} & \faFemale & \faMale & \textbf{\#} & \faClock & \faImages & \faHourglassHalf & \faMale & \faGlasses & \faTree & \faBinoculars & \faVideo & \faBullseye & \faHome & \faCloudSun \\
    \midrule  
        
        \multicolumn{17}{l}{\textbf{Complete Dataset}} \\
    \midrule
        Office Bldg. & L1 & St & & \faMale & 2 & 0.71 & 1.27 & 0.35 & \cm & \cm & \cm &  & & & \cm & \\
        
        \rowcolor{Gray}
        Robotics Hall & X1 & P & & \faMale & 6 & 2.50 & 4.50 & 0.42 & \cm & \cm & \cm & & & &  \cm & \\
        \rowcolor{Gray}
         & X2 & S,P & & \faMale & 6 & 19.00 & 34.20 & 3.17 & \cm & \cm & \cm & & & \cm & \cm & \\
        \rowcolor{Gray}
         & X3 & S,P & \faFemale\faFemale & \faMale\faMale & 9 & 23.29 & 41.93 & 2.59 & \cm & \cm & \cm & \cm & \cm & \cm &  \cm & \\
        Gym Hall & G1 & St,C,P,F & \faFemale\faFemale & \faMale & 8 & 10.24 & 18.44 & 1.28 & \cm & \cm & \cm & (\cm) & \cm & & \cm & \\
         & G2 & P,F & & \faMale & 4 & 3.23 & 5.82 & 0.81 & \cm & \cm & \cm & \cm & \cm & & \cm & \\
        \rowcolor{Gray}
        Debris Field & B1 & St,P & & \faMale\faMale\faMale\faMale & 20 & 23.01 & 41.42 & 1.15 & \cm & \cm & \cm & \cm & \cm & & & \cm\\
    
    \midrule
    \addlinespace[1ex]
        \multicolumn{17}{l}{\textbf{Test Subset}} \\
    \midrule
        Robotics Hall & X2 & S,P &  & \faMale & 6 & 19.00 & 34.20 & 3.17 & \cm & \cm & \cm & & \cm & \cm & \cm & \\
         & X3 & S,P & \faFemale\faFemale & \faMale\faMale & 9 & 23.29 & 41.93 & 2.59 & \cm & \cm & \cm & \cm & \cm & \cm & \cm & \\
    
    \midrule
    \addlinespace[1ex]
        \multicolumn{17}{l}{\textbf{Multi-View Subset (2nd Aria Observer)}} \\
    \midrule
        Robotics Hall & X3 & S,P,C & \faFemale\faFemale & \faMale\faMale & 9 & 23.29 & 41.93 & 2.59 & \cm & \cm & \cm & \cm & \cm & \cm & \cm & \\
        
        \rowcolor{Gray}
        Gym Hall & G1 & St,C,P,F & \faFemale\faFemale & \faMale & 3 & 3.54 & 6.37 & 1.18 & \cm & \cm & \cm & \cm & \cm & & \cm & \\
        \rowcolor{Gray}
         & G2 & P,F & & \faMale & 4 & 3.23 & 5.82 & 0.81 & \cm & \cm & \cm & \cm & \cm & & \cm & \\
         
        Debris Field & B1 & St,P & & \faMale\faMale\faMale\faMale & 20 & 23.01 & 41.42 & 1.15 & \cm & \cm & \cm & \cm & \cm & & & \cm \\
    
    \bottomrule
    \end{tabular}
    \vspace{1.5ex}
    \begin{tablenotes}[flushleft]
    \scriptsize 
    \item[] \textbf{Action:} \textbf{S:} Sitting/Lay, \textbf{St:} Stairs, \textbf{P:} Parkour, \textbf{C:} Climbing, \textbf{F:} Flips.
    \item[] \textbf{Sequence:} \textbf{\#} \#Seq., {\faClock}~Tot. Dur (min), {\faImages}~Tot. Frames (k), {\faHourglassHalf}~Avg. Len (min).
    \item[] \textbf{Sensors:} {\faGlasses}~Aria, {\faMale}~Rokoko, {\faTree}~BLK2GO, {\faBinoculars}~2nd Aria, {\faVideo}~Go-Pro, {\faBullseye}~Qualisys.
    \item[] \textbf{Location:} {\faHome}~Indoor, {\faCloudSun}~Outdoor. 
    \item[] (\cm) partial availability.
    \end{tablenotes}
    
\end{threeparttable}
\end{table}

\clearpage
\begin{table}[htbp]
    \centering
    
    \label{tab:app_sequence_overview}
    \begin{tabular}{l l l c c c}
        \toprule
        \textbf{Sequence} & \textbf{Subject} & \textbf{Location} & \textbf{2nd Aria} & \textbf{Pseudo GT} & \textbf{Duration (s)} \\
        \midrule
        cliff-jump\_s4\_bzb\_01 & s4 & B1 & \cm & $\times$ & 101.27 \\
        cliff-jump\_s4\_bzb\_02 & s4 & B1 & \cm & $\times$ & 80.17 \\
        cliff-jump\_s7\_bzb\_01 & s7 & B1 & \cm & $\times$ & 46.97 \\
        cliff-jump\_s8\_bzb\_01 & s8 & B1 & \cm & $\times$ & 115.77 \\
        cliff\_s4\_bzb\_01 & s4 & B1 & \cm & $\times$ & 64.77 \\
        crawling\_s1\_bzb\_02 & s1 & B1 & \cm & $\times$ & 88.13 \\
        crawling\_s1\_bzb\_03 & s1 & B1 & \cm & $\times$ & 96.23 \\
        rescue\_s4\_bzb\_01 & s4 & B1 & \cm & $\times$ & 62.50 \\
        rescue\_s8\_bzb\_01 & s8 & B1 & \cm & $\times$ & 70.93 \\
        rocky-path\_s7\_bzb\_01 & s7 & B1 & \cm & $\times$ & 64.17 \\
        rocky-path\_s7\_bzb\_02 & s7 & B1 & \cm & $\times$ & 55.77 \\
        rocky-path\_s8\_bzb\_01 & s8 & B1 & \cm & $\times$ & 86.47 \\
        walking\_rocky\_s7\_bzb\_01 & s7 & B1 & \cm & $\times$ & 77.37 \\
        walking\_rocky\_s7\_bzb\_02 & s7 & B1 & \cm & $\times$ & 72.57 \\
        walking\_s7\_bzb\_01 & s7 & B1 & \cm & $\times$ & 70.97 \\
        window-rocky\_s7\_bzb\_01 & s7 & B1 & \cm & $\times$ & 66.00 \\
        window-rocky\_s7\_bzb\_02 & s7 & B1 & \cm & $\times$ & 44.73 \\
        window\_s4\_bzb\_01 & s4 & B1 & \cm & $\times$ & 60.80 \\
        window\_s8\_bzb\_01 & s8 & B1 & \cm & $\times$ & 47.97 \\
        climbing\_s3\_gym\_01 & s3 & G1 & $\times$ & $\times$ & 105.90 \\
        crawling\_s3\_gym\_01 & s3 & G1 & $\times$ & $\times$ & 69.30 \\
        double-high-object\_s2\_gym\_01 & s2 & G1 & $\times$ & $\times$ & 78.37 \\
        overhang-object\_s2\_gym\_03 & s2 & G1 & $\times$ & $\times$ & 37.83 \\
        parkour\_s1\_gym\_02 & s1 & G1 & \cm & $\times$ & 45.33 \\
        parkour\_s1\_gym\_03 & s1 & G1 & \cm & $\times$ & 62.27 \\
        parkour\_s3\_gym\_01 & s3 & G1 & $\times$ & $\times$ & 110.77 \\
        walking\_s2\_gym\_01 & s2 & G1 & \cm & $\times$ & 104.77 \\
        parkour\_s1\_gym\_04 & s1 & G2 & \cm & $\times$ & 41.10 \\
        parkour\_s1\_gym\_05 & s1 & G2 & \cm & $\times$ & 45.03 \\
        parkour\_s1\_gym\_06 & s1 & G2 & \cm & $\times$ & 66.10 \\
        parkour\_s1\_gym\_07 & s1 & G2 & \cm & $\times$ & 41.67 \\
        double-stairs\_lee\_01 & s1 & L1 & $\times$ & $\times$ & 20.07 \\
        double-stairs\_lee\_02 & s1 & L1 & $\times$ & $\times$ & 22.37 \\
        cliff-path\_xhall\_01 & s1 & X1 & $\times$ & $\times$ & 29.50 \\
        crawling\_xhall\_01 & s1 & X1 & $\times$ & $\times$ & 23.23 \\
        double-high-object\_xhall\_01 & s1 & X1 & $\times$ & $\times$ & 28.60 \\
        high-object\_xhall\_01 & s1 & X1 & $\times$ & $\times$ & 22.87 \\
        narrow-path\_xhall\_01 & s1 & X1 & $\times$ & $\times$ & 22.43 \\
        stepping-stones\_xhall\_01 & s1 & X1 & $\times$ & $\times$ & 23.23 \\
        exomc-parkour\_xhall\_01 & s1 & X2 & $\times$ & \cm & 15.00 \\
        exomc-parkour\_xhall\_02 & s1 & X2 & $\times$ & \cm & 296.50 \\
        exomc-parkour\_xhall\_03 & s1 & X2 & $\times$ & \cm & 72.43 \\
        exomc-parkour\_xhall\_04 & s1 & X2 & $\times$ & \cm & 153.07 \\
        exomc-parkour\_xhall\_05 & s1 & X2 & $\times$ & \cm & 304.90 \\
        exomc-parkour\_xhall\_06 & s1 & X2 & $\times$ & \cm & 298.00 \\
        exo-cliff-stones\_s4\_xhall\_01 & s4 & X3 & \cm & \cm & 130.00 \\
        exo-cliff-stones\_s5\_xhall\_01 & s5 & X3 & \cm & \cm & 244.10 \\
        exo-high-object\_s4\_xhall\_01 & s4 & X3 & \cm & \cm & 130.00 \\
        exo-high-object\_s5\_xhall\_01 & s5 & X3 & \cm & \cm & 130.00 \\
        exo-parkour\_s4\_xhall\_01 & s4 & X3 & \cm & \cm & 130.00 \\
        exo-standing\_s1\_xhall\_01 & s1 & X3 & \cm & \cm & 130.00 \\
        exo-standing\_s6\_xhall\_01 & s6 & X3 & \cm & \cm & 243.53 \\
        exo-walking\_s4\_xhall\_01 & s4 & X3 & \cm & \cm & 130.00 \\
        exo-walking\_s5\_xhall\_01 & s5 & X3 & \cm & \cm & 130.00 \\
        \bottomrule
    \end{tabular}
    \vspace{6pt}
    \captionsetup{labelfont=bf, font=small, skip=0pt, justification=justified}
    \caption{Overview of sequence locations, subjects, sensor modalities, and durations.}
\end{table}
\clearpage

\section{Experiments}\label{sec:app_exp}

\subsection{Extended Quantitative Dataset Evaluation} 
\label{sec:app_quantitative_dataset_eval}
We expand upon the quantitative evaluation of our dataset by detailing the specific metrics used for local and global pose estimation, while also clarifying the differing evaluation protocols of prior baselines to ensure a fair contextual comparison.

We evaluated the local and global human pose and shape estimation on our test set (reported in \cref{tab:dataset}) on the acquired motion capture ground truth, described in \cref{sec:gt_acquisition}. 
Following standard evaluation protocols in recent literature, we assess local pose estimation using mean per-joint position error (MPJPE) and Procrustes-aligned MPJPE (PA-MPJPE) \citep{PROX2019, RICHHuang:CVPR:2022, Dai_2023_CVPR, wang2025prompthmrpromptablehumanmesh, multi-hmr2024}. 
To evaluate global tracking accuracy, we report world MPJPE (W-MPJPE) and world Procrustes-aligned MPJPE (WA-MPJPE) \citep{wang2024tram, wham:cvpr:2024, liu2026joint, chen2025human3r}, aligning ground truth and test set joint positions of either the first two frames or all frames of 100-frame segments. We report all error metrics in millimeters. 
We evaluate the root translation error (RTE), normalized over the total accumulated error.   
Foot sliding (FS) is the average foot displacement between adjacent
frames where the ground truth foot displacement is less than a threshold (10mm). 
Finally, we evaluate the foot floating rate (FFR). The rate computes the minimum distance between the human foot and the scene point cloud and reports the ratio of the distance greater than a given threshold (10cm).

To contextualize our performance, we evaluate local and global Human Pose and Shape (HPS) estimation on our test set and compare the results against reported state-of-the-art baselines (Table 2). However, directly comparing these numbers requires careful consideration, as each prior work employs a distinct quantitative evaluation protocol.\\
PROX \citep{PROX2019} evaluates its test set directly against MoCap ground truth test set. While they also propose a PROX-D variant that achieves superior results using privileged depth data, we exclude it here to ensure a fair comparison. Further, we derive the HPS dataset \citep{Guzov2021HPS} metrics by averaging the errors across the quantitatively evaluated dynamic motion sequences. RICH \citep{RICHHuang:CVPR:2022} evaluates human pose and shape on regressed SMPL-X bodies \citep{PIXIE:3DV:2021}, explicitly differentiating between frames with and without complex scene contacts (i.e., non-foot-ground interactions). We compare against their superior, non-contact frames. Finally, because SLOPER4D \citep{Dai_2023_CVPR} lacks direct ground-truth data, the authors perform a modality-separated (camera and LiDAR) cross-evaluation on different datasets. To provide a consolidated baseline in Table 2, we report the mean of their best-performing results.

\subsection{Human Pose Accuracy}
An extended subset evaluation (\cref{fig:motion_accuracy}) confirms that the local pose and RTE metrics remain consistent across subjects (listed in \cref{tab:subject_measurements_id_sorted}). The world-aligned distance error (WA-MPJPE) indicates robust underlying motion trajectories across the test set, while the absolute global positioning (W-MPJPE) shows subject-specific variance. The difference between subject 1 and 6 can be traced back to differences in the tested motion patterns (compare with \cref{tab:app_sequence_overview}).

\begin{table}[h]
\centering
\captionsetup{labelfont=bf, font=small, skip=10pt, justification=justified}
\caption{Anthropometric Measurements for EgoHTR Subjects}
\label{tab:subject_measurements_id_sorted}
\scriptsize
\setlength{\tabcolsep}{3.5pt}
\renewcommand{\arraystretch}{1.2}
\setlength{\aboverulesep}{1pt}
\setlength{\belowrulesep}{2pt}
\begin{tabular}{l | c | c | c | c | c | c | c | c}
    \toprule
    \textbf{Measurement Item} & \textbf{Subj. 1} & \textbf{Subj. 2} & \textbf{Subj. 3} & \textbf{Subj. 4} & \textbf{Subj. 5} & \textbf{Subj. 6} & \textbf{Subj. 7} & \textbf{Subj. 8} \\
    \midrule
    \textbf{Mass (kg)}        & 75.0             & 75.0             & 72.0             & 68.0             & 65.0             & 50.0             & 73.0             & 73.0             \\
    \textbf{Total Height (cm)} & 180.0            & 175.0            & 162.0            & 170.0            & 170.0            & 160.0            & 184.0            & 178.0            \\
    \midrule
    Arm Span (cm)             & 192.2            & 182.0            & 170.0            & 181.5            & 181.5            & 157.0            & 185.0            & 176.0            \\
    Shoulder Width (cm)       & 41.7             & 35.0             & 38.0             & 39.4             & 32.0             & 31.0             & 37.0             & 38.2             \\
    Shoulder Height (cm)      & 149.5            & 145.3            & 133.0            & 141.2            & 141.2            & 129.0            & 151.0            & 147.0            \\
    Hip Width (cm)            & 22.8             & 26.0             & 25.0             & 21.5             & 21.5             & 25.0             & 25.0             & 22.5             \\
    Hip Height (cm)           & 103.7            & 100.8            & 92.3             & 98.0             & 100.0            & 85.0             & 102.0            & 105.0            \\
    Knee Height (cm)          & 51.7             & 50.2             & 48.0             & 48.8             & 49.8             & 42.0             & 50.8             & 52.3             \\
    Manus Length (cm)         & 48.7             & 47.0             & 44.0             & 46.0             & 44.5             & 41.2             & 46.0             & 45.2             \\
    Hand Length (cm)          & 19.5             & 18.8             & 19.0             & 18.4             & 17.8             & 17.0             & 18.4             & 19.0             \\
    Hand Width (cm)           & 8.7              & 8.4              & 8.5              & 8.2              & 7.9              & 7.6              & 8.2              & 8.5              \\
    Foot Length (cm)          & 26.0             & 26.0             & 25.0             & 24.6             & 24.6             & 22.0             & 26.6             & 25.7             \\
    \bottomrule
\end{tabular}
\end{table}

\begin{figure}[h] \label{tab:subjects}
    \centering
    \captionsetup{labelfont=bf, font=small, skip=10pt, justification=justified}
    \caption{Subset evaluation of the local and global body pose estimation of the EgoHTR test set on the MoCap ground truth. The subsets illustrate the results of the participating male (s1, s2) and female (s4, s5) subjects.}
    \includegraphics[width=\textwidth]{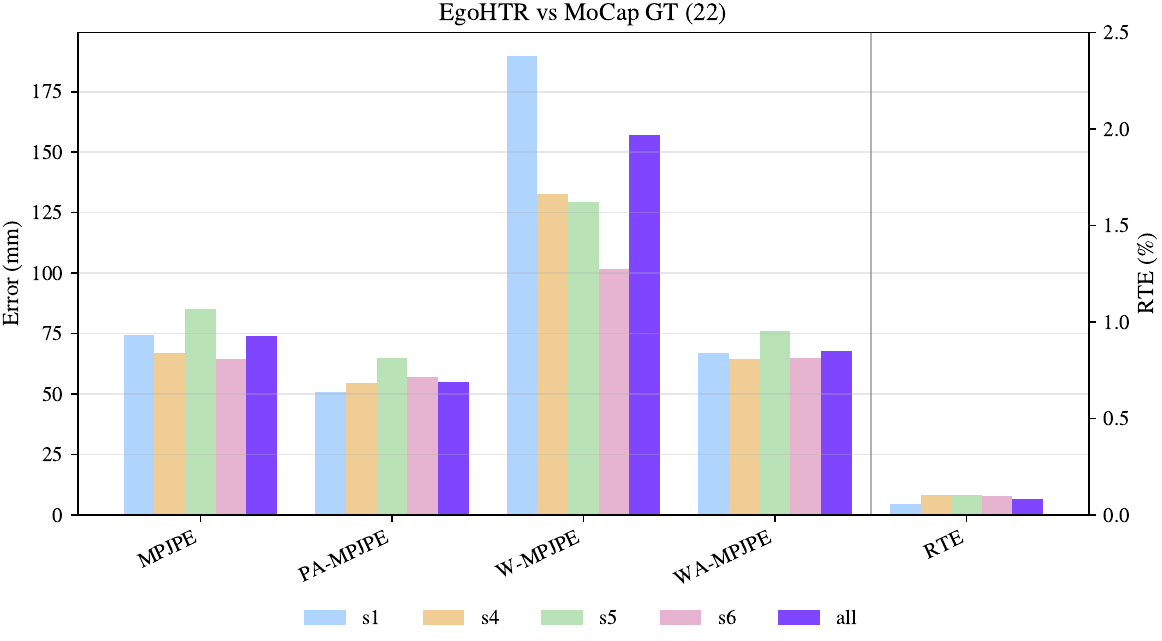}
    \label{fig:motion_accuracy}

    \scriptsize
    \setlength{\tabcolsep}{4pt} 
    \renewcommand{\arraystretch}{1.2} 
    \centering

    \begin{tabular}{l cc cc ccc}
        \toprule
        & \multicolumn{2}{c}{} 
        & \multicolumn{2}{c}{\textbf{Local}} 
        & \multicolumn{3}{c}{\textbf{Global}} \\
        
        \cmidrule(lr){4-5} \cmidrule(lr){6-8} 
        
         & Sequences & Frames & MPJPE $\downarrow$ & PA-MPJPE $\downarrow$ & WA-MPJPE $\downarrow$ & W-MPJPE $\downarrow$ & RTE $\downarrow$ (\%) \\
        \midrule
        
        
        
        \rowcolor{gray!10} 
        \textbf{EgoHTR (Test)} & 15 & 72.6k & 73.2 & 54.3 & 66.7 & 151.3 & 0.09 \\
        
        \hspace{2mm} Subject 1 & 7 & 37.3k & 73.0 & \textbf{49.8} & \textbf{64.5} & 175.3 & \textbf{0.09} \\
        \hspace{2mm} Subject 4 & 4 & 15.5k & \underline{66.8} & \underline{54.5} & \underline{64.6} & 132.5 & \underline{0.10} \\
        \hspace{2mm} Subject 5 & 3 & 13.6k & 85.3 & 65.1 & 76.0 & \underline{129.4} & 0.11 \\
        \hspace{2mm} Subject 6 & 1 & 6.2k  & \textbf{64.4} & 57.2 & 64.8 & \textbf{101.9} & \underline{0.10} \\
        
        \bottomrule
        
    \end{tabular}
    \vspace{6pt}
    \captionsetup{labelfont=bf, font=small, skip=4pt, justification=justified}
    \captionof{table}{Exact numerical values for the human pose accuracy evaluation, corresponding to the results visualized in Figure \ref{fig:motion_accuracy}. Best values in \textbf{bold}, second best \underline{underlined}.}
    \label{tab:subjects}
    
\end{figure}

\clearpage
\subsubsection{Motion Capture Ground Truth Verification}

\begin{figure}[htbp]
    \centering
    \captionsetup{labelfont=bf, font=small, skip=15pt, justification=justified}
    \caption{Verification of the MoCap ground truth on local and global pose estimation. Two baselines of different modalities, HPS estimation from egocentric camera and 4D human-scene reconstruction, are evaluated on our test set and MoCap ground truth.}
    \includegraphics[width=\textwidth]{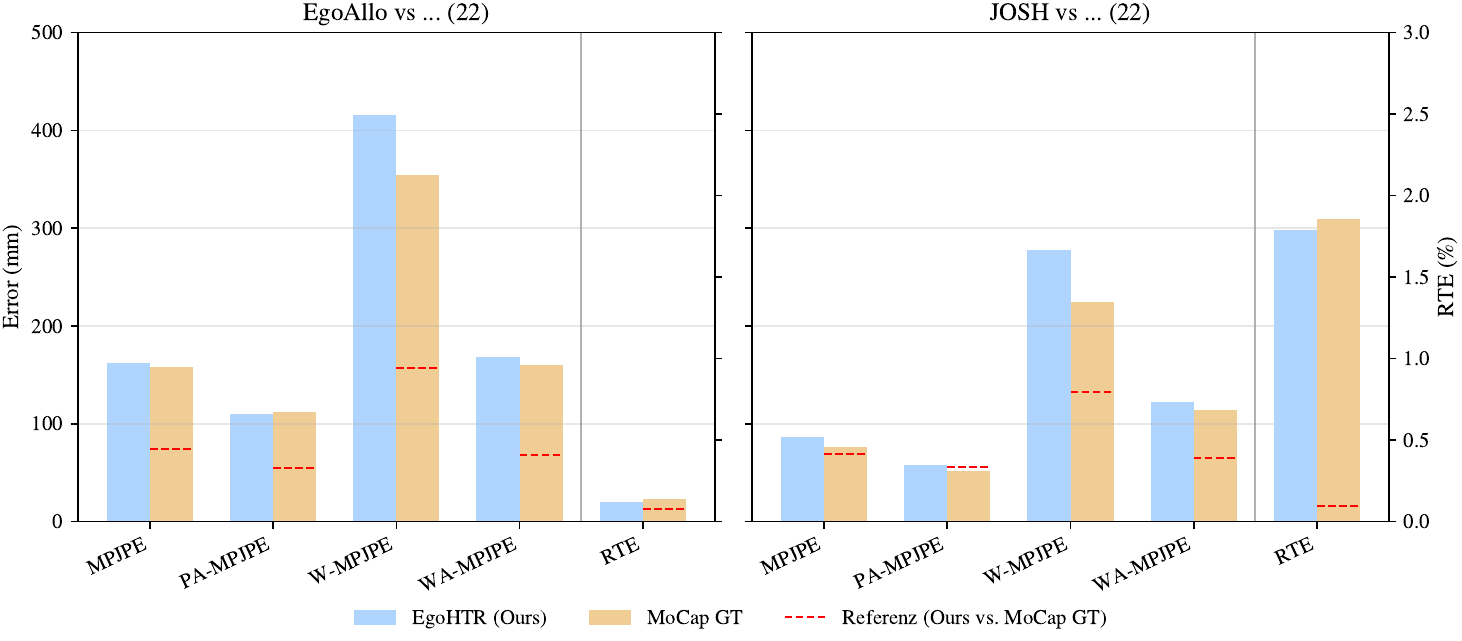}
    \label{fig:gt_verification}
    
    \scriptsize 
    \setlength{\tabcolsep}{4pt} 
    \renewcommand{\arraystretch}{1.1} 
    
    \begin{tabular}{l c cc ccc c c c cc ccc}
        \toprule
        
        & \multicolumn{6}{c}{\textbf{EgoAllo (22)}} 
        &&& \multicolumn{6}{c}{\textbf{JOSH (22)}} \\
        
        \cmidrule(lr){2-7} \cmidrule(lr){10-15}
        
        & 
        & \multicolumn{2}{c}{\textbf{Local}} 
        & \multicolumn{3}{c}{\textbf{Global}}
        & 
        & 
        &
        & \multicolumn{2}{c}{\textbf{Local}} 
        & \multicolumn{3}{c}{\textbf{Global}} \\
        
        \cmidrule(lr){3-4} \cmidrule(lr){5-7} \cmidrule(lr){11-12} \cmidrule(lr){13-15}
        
        \textbf{Evaluated on ...}
        & \rotatebox{90}{\tiny  Frames (k)}
        & \rotatebox{90}{\tiny   MPJPE $\downarrow$}
        & \rotatebox{90}{\tiny   PA-MPJPE $\downarrow$}
        & \rotatebox{90}{\tiny   W-MPJPE $\downarrow$}
        & \rotatebox{90}{\tiny   WA-MPJPE $\downarrow$}
        & \rotatebox{90}{\tiny   RTE (\%) $\downarrow$}
        &&& 
        \rotatebox{90}{\tiny   Frames (k)} 
        & \rotatebox{90}{\tiny   MPJPE $\downarrow$}
        & \rotatebox{90}{\tiny   PA-MPJPE $\downarrow$}
        & \rotatebox{90}{\tiny   W-MPJPE $\downarrow$}
        & \rotatebox{90}{\tiny   WA-MPJPE $\downarrow$}
        & \rotatebox{90}{\tiny   RTE (\%) $\downarrow$} \\
        
        \midrule
        
        EgoHTR (Ours)  & 83.7 & 161.8 & 110.2 & 405.5 & 167.1 & 0.13 &&& 10.5 & 89.4 & 60.3 & 212.3 & 104.9 & 2.45 \\
        MoCap GT   & 72.3 & 158.2 & 111.6 & 346.2 & 159.3 & 0.15 &&& 10.01 & 75.3 & 52.8 & 175.3 & 95.1  & 2.59 \\
        \midrule
        Ref (Ours vs MoCap GT) & 72.6 & 73.2 & 54.3 & 66.7 & 151.3 & 0.09 &&& 22.1 & 70.3 & 56.4 & 129.4 & 66.2  & 0.11 \\
        
        \bottomrule
    \end{tabular}
    \centering
    \captionsetup{labelfont=bf, font=small, skip=15pt, justification=justified}
    \captionof{table}{Exact numerical values for the ground truth verification, corresponding to the results visualized in Figure \ref{fig:gt_verification}. EgoAllo \citep{yi2025egoallo} and JOSH \citep{liu2026joint} are evaluated on 15 sequences and 6 sequences, respectively} 
    \label{tab:gt_verification}
    
\end{figure}

To validate the quality of our proposed test set, we compared the performance of two state-of-the-art baselines, EgoAllo \citep{yi2025egoallo} and JOSH \citep{liu2026joint}, when evaluated on EgoHTR versus the motion capture ground truth (MoCap GT). As illustrated in \cref{fig:gt_verification} and detailed in \cref{tab:gt_verification}, the local metrics (MPJPE and PA-MPJPE) show marginal differences when evaluating EgoAllo or JOSH on EgoHTR compared to the evaluation on MoCap GT (\SI{3.6}{mm}/\SI{14.1}{mm} for EgoAllo and \SI{1.4}{mm}/\SI{7.5}{mm} for JOSH, respectively). Similarly, global trajectory tracking (RTE) differences remain small at just \SI{0.02}{\%} and \SI{0.14}{\%}, respectively. The larger offset for the global distance metrics has two reasons: Overall, the gap between the test set and the MoCap GT correlates with the absolute magnitude of the measured error. Therefore, the larger discrepancies seen in global spatial distance error are expected. Additionally, the W-MPJPE captures the non-negligible jitter present in the reconstructed body pose.

The consistency across metrics confirms that EgoHTR provides a robust approximation of the motion capture ground truth, validating its use as a reliable evaluation baseline.

\clearpage

\subsection{Perceptive Locomotion from Demonstrations}
\subsubsection{Training pipeline}\label{app:training_pipeline} 
We train per-clip expert policies with PPO~\citep{ppo} in Isaac Lab,
instantiating a Unitree G1 humanoid on the terrain mesh paired with each
reference motion. The actor observes proprioception, the reference joint
command, and a yaw-aligned height scan; the critic additionally receives
privileged world-frame body states and ankle contact wrenches. Actions
are joint-position targets for all 29 DoFs. Rewards combine the
world-frame anchor and feet/hand position trackers, relative body-pose
and velocity tracking, the foot-contact reward and standard
regularizer (action rate, joint limits, undesired contacts). Episodes
are initialized by adaptive reference state
initialization~\citep{DeepMimic, liao2025beyondmimicmotiontrackingversatile}
restricted to frames free of robot--terrain interpenetration, and
terminated on anchor or end-effector drift. For hardware deployment we
use the same pipeline with added observation noise, actuator-gain and
friction randomization, base-mass perturbations, and foot wrench
disturbances to emulate shaking of obstacles that are not fixed to the ground. Additionally, we remove the base linear velocity observation, as it cannot be reliably estimated on the real hardware (without global tracking systems).

\begin{figure}[htbp]
    \vspace{-5pt}
    \centering
    \includegraphics[trim={0cm 0cm 0cm 0cm}, clip, width=\textwidth]{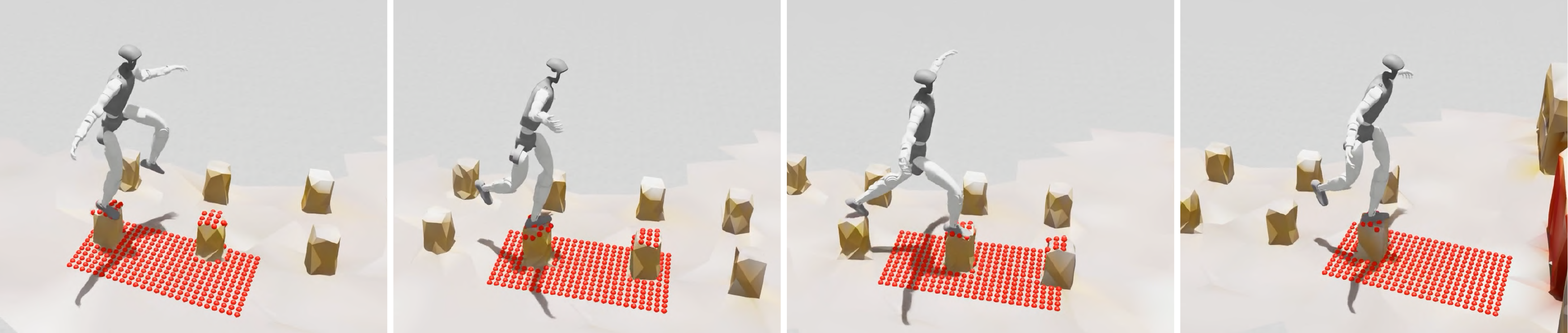}
    \captionsetup{labelfont=bf, font=small, skip=5pt, justification=justified}
    \caption{Deployment of stepping stones motion tracking policy in simulation.}
    \label{app:stepping_stones}
    \vspace{-5pt}
\end{figure}

\begin{figure}[htbp]
    \vspace{-5pt}
    \centering
    \includegraphics[trim={0cm 0cm 0cm 0cm}, clip, width=\textwidth]{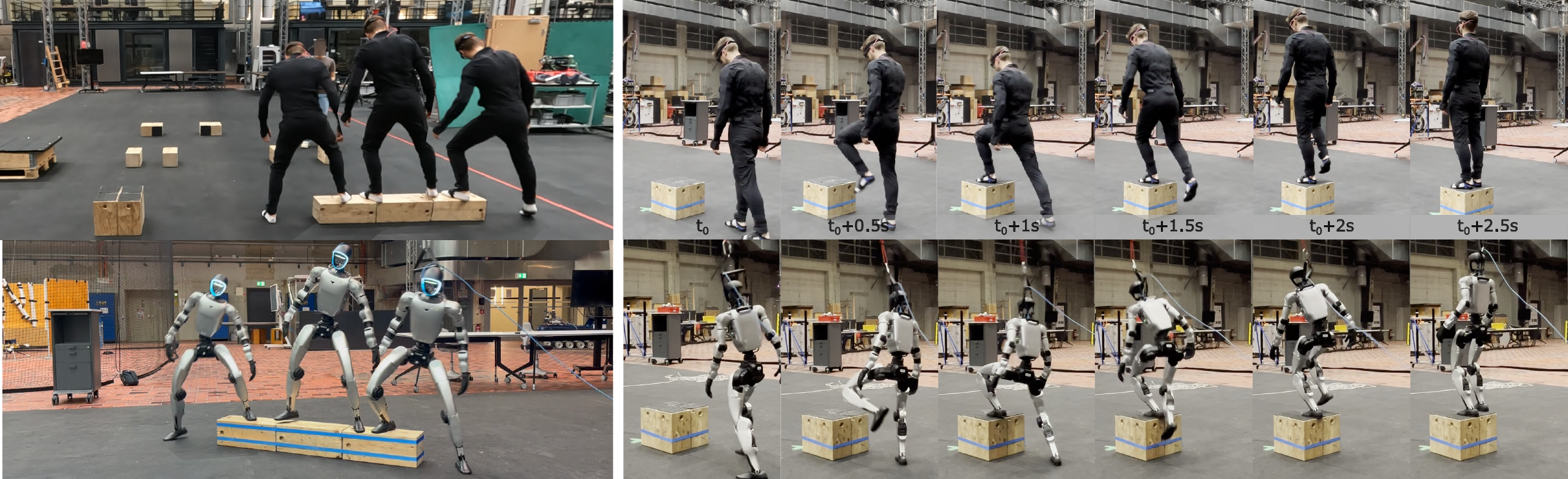}
    \captionsetup{labelfont=bf, font=small, skip=5pt, justification=justified}
    \caption{Hardware deployment of perceptive policies trained on an atomic beam (left) and box-up motion (right) from EgoHTR, executed on a Unitree G1.}
    \label{app:deploy_hardware}
    \vspace{-5pt}
\end{figure}

\subsection{Human Motion Generation} \label{app:human_motion_generation}
The application of our dataset can be extended to human motion synthesis.
Existing motion generators are typically conditioned on simplified spatial priors, such as 2D trajectories, basic obstacle avoidance heuristics and sparse spatial key frames \cite{karunratanakul2023guided, huang2025diffuse, rempe2026kimodo} or gaze-informed human motion with global scene features \citep{zheng2022gimo}. 
While sufficient for flat surfaces, these sparse conditioning signals fail entirely on unstructured environment, where continuous, high-resolution environmental geometry must dictate precise foot placement. Overcoming this limitation requires training on high-fidelity human-scene interaction data with local geometric precision.


\end{document}